\pdfoutput=1
\PassOptionsToPackage{dvipsnames,svgnames,table}{xcolor}
\documentclass[11pt]{article}
\usepackage[]{acl}

\usepackage{hyperref}
\usepackage{adjustbox}
\usepackage{graphicx}
\usepackage{enumitem}
\usepackage{multicol}
\usepackage{amsmath}
\usepackage{bold-extra}
\usepackage{subcaption}
\usepackage{float}
\usepackage{nicefrac}
\usepackage{color, colortbl}
\usepackage{times}
\usepackage{latexsym}
\usepackage{makecell}
\usepackage{amssymb}
\usepackage{multirow}
\usepackage{tablefootnote}
\usepackage{textcomp}
\usepackage{CJKutf8}
\usepackage{booktabs}
\usepackage{siunitx}
\usepackage{authblk}
\usepackage{tikz}
\usepackage[T1]{fontenc}

\usepackage[utf8]{inputenc}

\usepackage{microtype}

\usepackage{inconsolata}
\title{Do Moral Judgment and Reasoning Capability of LLMs Change with Language? A Study using the Multilingual Defining Issues Test}

\author{ {\bf Aditi Khandelwal}{$^1$}\thanks{\enspace Equal contribution}\hspace{0.4cm}
{\bf Utkarsh Agarwal}{$^1$}\footnote[1]\hspace{0.4cm}
{\bf Kumar Tanmay}{$^1$}\footnote[1]\hspace{0.4cm}
{\bf Monojit Choudhury}{$^2$}\thanks{\enspace Work done while at Microsoft.}\\
\textsuperscript{$1$}Microsoft Corporation\\
\textsuperscript{$2$}MBZUAI\\
{\{t-aditikh, t-utagarwal, t-ktanmay\}@microsoft.com, monojit.choudhury@mbzuai.ac.ae}
}

\begin{document}
\maketitle
\begin{abstract}
This paper explores the moral judgment and moral reasoning abilities exhibited by Large Language Models (LLMs) across languages through the Defining Issues Test. It iswe a well known fact that moral judgment depends on the language in which the question is asked~\cite{morals_language}. We extend the work of \citet{tanmay2023probing} beyond English, to 5 new languages (Chinese, Hindi, Russian, Spanish and Swahili), and probe three LLMs -- ChatGPT, GPT-4 and Llama2Chat-70B -- that shows substantial multilingual text processing and generation abilities. Our study shows that the moral reasoning ability for all models, as indicated by the post-conventional score, is substantially inferior for Hindi and Swahili, compared to Spanish, Russian, Chinese and English, while there is no clear trend for the performance of the latter four languages. The moral judgments too vary considerably by the language.
\end{abstract}

\section{Introduction}

In a recent work, \citet{tanmay2023probing} used the Defining Issues Test (DIT) \cite{Rest1986B}, a psychological assessment tool based on Kohlberg’s Cognitive Moral Development (CMD) \cite{KohlbergCMD}, to evaluate the moral reasoning capabilities of large language models (LLMs) such as GPT-4, ChatGPT, Llama2Chat-70B and PaLM-2. The DIT presents a moral dilemma along with 12 statements on ethical considerations and asks the respondent (in our case, the LLM) to rank them in order of importance for resolving the dilemma. The test outcome is a set of scores that indicate the respondent’s moral development stage. According to this study~\cite{tanmay2023probing}, GPT-4 was found to have the best moral reasoning capability, equivalent to that of a graduate student, while the three other models exhibited a moral reasoning ability that is at par with an average adult. 

Although interesting, the study was limited to English, even though many of the models studied were multilingual. On the other hand, it is known that, for humans, moral judgment often depends on the language in which the dilemma is presented~\cite{morals_language}. Language is a powerful tool that shapes our thoughts, beliefs and actions. It can also affect how we perceive and resolve moral dilemmas. Research in moral psychology has shown that people are more likely to endorse utilitarian choices (such as sacrificing one person to save five) when they read a dilemma in a foreign language (L2) than in their native language (L1) \cite{circi2021foreign, corey2017our}. This suggests that language can modulate our emotional and cognitive responses to moral situations.

To what extent does the moral judgment and reasoning capability of LLMs depend on the language in which the question is asked, and what are the factors responsible for the differences across languages, if any? In this paper, we extend the DIT-based study by~\citet{tanmay2023probing} to five languages -- Spanish, Russian, Chinese, Hindi and Swahili. We study three popular LLMs - GPT-4 \cite{openai2023gpt4}, ChatGPT \cite{chatgpt} and Llama2Chat-70B \cite{touvron2023llama}, by probing them with the dilemmas and the moral considerations separately for each language. We prompt the model to provide a resolution to the dilemma and the list of top 4 most important moral considerations. The responses are then used to compute the moral staging scores of the LLMs for different languages. 

Some of the salient observations of this study are: (1) GPT-4 has the best multilingual moral reasoning capability with minimal difference in moral judgment and staging scores across languages, while for LLama2Chat-70B and ChatGPT the performance varies widely; (2) For all models, we observe superior moral reasoning abilities for English and Spanish followed by Russian, Chinese, Swahili and Hindi (in descending order of performance). Performance in Hindi for ChatGPT and LLama2Chat-70B is no better than a random baseline. (3) Despite high moral staging score for both English and Russian, we find significant differences in moral judgment for these two languages, while the judgments for English, Chinese and Spanish tend to agree more often. 

While the difference in moral reasoning abilities across languages seem correlated to the amount of resources available or used for training the models, the reason behind the differences and similarities in the moral judgments across the high resource languages (i.e., Chinese, English, Russian and Spanish) is not obvious. We speculate it to be reflective of the values of the societies where these languages are spoken, but also propose alternative hypotheses.

Apart from being the first multilingual study of moral reasoning ability of LLMs in the framework of Kohlberg's CMD model, one key contribution of this work is the creation of multilingual versions of the moral dilemmas presented in DIT~\cite{Rest1986B} and \citet{tanmay2023probing}. We will publicly share these datasets, subject to permissions from the original authors.

\section{Background: Moral Psychology and Ethics of NLP}

{\em Morality}, the study of right and wrong, has long been a central topic in philosophy \cite{gert2002definition}. The Cognitive Moral Development (CMD) model by Lawrence Kohlberg \citeyear{kohlberg1981philosophy} is a prominent theory that categorizes moral development into three levels: {\em pre-conventional}, {\em conventional}, and {\em post-conventional} morality. The Defining Issues Test (DIT) by James Rest \citeyear{rest1979development} measures moral reasoning abilities using moral dilemmas, providing insights into ethical decision-making.  This tool has been widely used for over three decades, providing insights into ethical decision-making processes \cite{rest1994moral}. 

\subsection{Defining Issues Test}

DIT consists of several moral dilemmas. As an illustration, consider {\bf Timmy's Dilemma}\footnote{DIT is behind a paywall, and hence, we cannot share the actual dilemmas publicly. Therefore, we use this dilemma proposed by~\citet{tanmay2023probing} as our running example}: Timmy is a software engineer, working on a crucial project that supports millions of customers. He discovers a bug in the deployed system, which, if not fixed immediately, could put the privacy of many customers at risk. Only Timmy knows about this bug and how to fix it. However, Timmy's best friend is getting married, and Timmy has promised to attend and officiate the ceremony. If he decides to fix the bug now, he will have to miss the wedding. Should Timmy go for the wedding (option 1), or fix the bug first (option 3)? Or maybe it is simply not possible to decide (option 2). 

In DIT, first, the respondent is asked to resolve such dilemmas that pit moral values (in Timmy's case between as professional vs. personal commitments) against each other. The resolution is called the {\em moral judgment} offered by the respondent. Then the respondent is presented with 12 {\em moral consideration} statements. For instance, ``Will Timmy get fired by his organization if he doesn't fix the bug?", or ``Should Timmy act according to his conscience and moral values of loyalty towards a friend, and attend the wedding?" They are asked to choose the 4 most important considerations (ranked by importance) that helped them arrive at the moral judgment. In other words, the respondent has to provide a {\em moral reasoning} for the judgment made. Each statement is assigned to a specific moral development stage of the CMD model. A set of moral development scores are then computed based on the response, which is explained in detail in Section~\ref{subsec:metrics}. Note that some statements are irrelevant or against the conventions of society, which are ignored during the analysis but can inform us about the attentiveness of the respondent.

\subsection{Moral Judgment vs. Moral Reasoning}
\label{subsec:moralreasoning}
There is a long standing debate in moral philosophy and psychology on what factors influence moral judgments~\cite{haidt2001emotional}. While prominent philosophers including Plato, Kant and Kohlberg have argued in favor of deductive reasoning (not necessarily limited to pure logic) as the underlying mechanism, recent research in psychology and neuroscience shows that in most cases people intuitively arrive at a moral judgment and then use post-hoc reasoning to rationalize it or explain/justify their position or to influence others in a social setting (see \citet{greene2002and} for a survey). In this sense, moral judgments are similar to aesthetic judgments rather than logical deductions. It also explains why policy-makers often decide in favor of wrong and unfair policies despite availability of clear evidence against those. 

Therefore, DIT as well as its very foundation, Kohlberg's CMD has been criticized for over-emphasis on moral reasoning over moral intuitions \cite{dien1982chinese, snarey1985cross,bebeau1987integrating,haidt2001emotional}. However, it will be interesting to test the moral intuition vs. reasoning hypothesis for LLMs, and what the alignment (or if we may say, ``moral intuition") of the popular models are~\cite{yao2023instructions}. 
However, it will not only be interesting to test the moral intuition vs. reasoning hypothesis for LLMs, and what the alignment (or if we may say, ``moral intuition") of the popular models are~\cite{yao2023instructions}, but also to explore how altering language affects moral reasoning capabilities, assuming the value of analyzing such reasoning has already been well argued for and established by previous work \cite{rao2023ethical}.

\subsection{Language and Morality}
Recent research \cite{morals_language, hayakawa2017thinking, corey2017our} reveals an intriguing connection between moral judgment and the "Foreign-Language Effect", that individuals tend to make more utilitarian choices when faced with moral dilemmas presented in a foreign language (L2), as opposed to their native tongue (L1). This shift appears to be linked to reduced emotional responsiveness when using a foreign language, leading to a diminished influence of emotions on moral judgments. \citet{vcavar2018moral} also shows how a higher proficiency and a higher degree of acculturation in L2 may reduce utilitarianism in the L2 condition. This suggests that linguistic factors can significantly influence moral decision-making, impacting a substantial number of individuals. There are more complex interactions among dilemma type, emotional arousal, and the language in bilingual individuals' moral decision making process \cite{chan2016effects}. 

\subsection{Current Approaches to Ethics of LLMs}

AI alignment aims to ensure AI systems align with human goals and ethics \cite{vox2018}. Several work provide ethical frameworks \cite{araque2020moralstrength}, guidelines, and datasets \cite{hoover2020moral, trager2022moral, alshomary2022moral} for training and evaluating LLMs in ethical considerations and societal norms \cite{ethicsdataset}. However, they may suffer from bias based on annotator backgrounds \cite{olteanu2019social}. Recent research emphasizes in-context learning and supervised tuning to align LLMs with ethical principles \cite{zhou2023rethinking, jiang2021can, rao2023ethical}. These methods accommodate diverse ethical views that are essential given the multifaceted nature of ethics. \citet{tanmay2023probing} introduce an ethical framework utilizing the Defining Issues Test to assess the ethical reasoning capabilities of LLMs. The authors assessed the models performance with moral dilemmas in English. To expand upon this work, our research delves deeper into the performance of these models when confronted with moral dilemmas in a multilingual context. This investigation aims to unveil how these LLMs respond to the same scenarios in different languages, shedding light on their cross-linguistic ethical reasoning capabilities.

\subsection{Performance of LLMs across Languages}
LLMs demonstrate impressive multilingual capability in natural language processing tasks, but their proficiency varies across languages \cite{zhao2023survey}. While their training data is primarily in English, it includes data from other languages \cite{brown2020language, chowdhery2022palm, zhang2022opt, zeng2022glm}. Despite their capabilities, the vast number of languages worldwide, most of which are low-resource, presents a challenge. LLMs still encounter difficulties with non-English languages, particularly in low-resource settings \cite{bang2023multitask, jiao2023chatgpt, hendy2023good, zhu2023multilingual}. Many studies have shown how the multilingual performances of the LLMs can be improved using in-context learning and carefully designed prompts \cite{huang2023not, nguyen2023democratizing}. \citet{ahuja2023mega} and \citet{wang2023seaeval} report experiments for benchmarking the multilingual capabilities of LLMs in various NLP tasks, such as Machine Translation, Natural Language Inference, Sentiment Analysis, Text Summarization, Named Entity Recognition, and Natural Language Generation, and conclude that LLMs do not perform well for most but a few high resource languages. \citet{kovavc2023large} show that LLMs exhibit varying context-dependent values and personality traits across perspectives, contrasting with humans, who typically maintain more consistent values and traits across contexts \cite{schwartz2012overview, graham2013moral}. 

Existing research on multilingual LLMs has primarily focused on technical capabilities, neglecting the exploration of their moral reasoning in diverse linguistic and cultural contexts. This underscores the importance of probing into the ethical dimensions of multilingual LLMs, given their significant impact on various real-life applications and domains.

\section{Experiments}
In this section, we provide an overview of our experimental setup, datasets, the language models (LLMs) that were studied, the structure of the prompts, and the metrics employed. Our prompts to the LLMs include a moral dilemma scenario, accompanied by a set of 12 ethical considerations and three subsequent questions. By analyzing the responses to these questions, we calculate the P-score as well as individual stage scores for each LLM which we explain in Section~\ref{subsec:metrics}.

\subsection{Dataset and Prompt}
We use the five dilemmas from DIT-1\footnote{Obtained the dataset by purchasing from The University of Alabama through the official website:  \url{https://ethicaldevelopment.ua.edu/ordering-information.html}} (Heinz, Newspaper, Webster, Student, Prisoner) and four dilemmas introduced by \citet{tanmay2023probing}. We used the Google Translation API to translate all these dilemmas into six different languages: Hindi, Spanish, Swahili, Russian, Chinese, and Arabic. To ensure the quality of translations, we asked native speakers of Swahili, Hindi, Chinese, Russian, and Spanish to verify them. They suggested some minor stylistic changes for 1-2 words per dilemma, but they confirmed that the meaning was preserved. We also back-translated them into English to check if the meaning remained consistent.  Our choice was guided by our aim to include diverse languages across three dimensions: (a) the amount of resource available -- Spanish, Chinese (high) to Hindi (medium) and Swahili (low); (b) the script used - Spanish and Swahili use the Latin script, while Hindi, Russian, Arabic, and Chinese employ non-Latin scripts, and (c) the cultural context of the L1 speakers of the languages -- Hindi and Swahili from Global South representing traditional value-based cultures, Russian for orthodox Europe, Spain for Catholic Europe and Chinese for Confucian system of values (based on World Value Survey by \citet{inglehart2010wvs}). We followed the same process as described in \citet{tanmay2023probing} for the prompt, translating it using the Google API and verifying the translations using the same technique mentioned above. The prompt structure can be found in Figure~\ref{fig:prompt} in the Appendix.

\subsection{Experimental Setup}

We examined three of the most prominent LLMs with multilingual capabilities \cite{wang2023seaeval}: GPT-4 (size undisclosed) \cite{openai2023gpt4}, ChatGPT with 175 billion parameters \cite{chatgpt}, and Llama2-Chat with 70 billion parameters \cite{touvron2023llama}. 
We applied the same shuffling strategy, again as described by \citet{tanmay2023probing}, in resolving dilemmas by selecting one of the three options (O1, O2, and O3) that is 6 permutations of options and considering 8 distinct permutations out of the possible 12 statements (out of 12! possibilities), resulting in a total of 48 permutations of prompts per dilemma per language. 

Throughout all our experiments, we set the temperature to 0, a presence penalty of 1, and a top probabilities value of 0.95. Furthermore, we specified a maximum token length of 2000 for English, Spanish, Chinese, Swahili, and Russian, while for Hindi, we set a maximum token length of 4000, as it requires a more tokens due to higher fertility of the tokenizer.

\subsection{Method}
We provide the translated prompt to the model and translate the response to English using Google Translate API. Then we extract the responses of the three questions posed in the DIT from the translated English response. We manually check the answers for quality and find that for Arabic, the responses for ChatGPT and Llama2Chat were getting truncated because of running out of maximum token length of 4000. So we had to leave out Arabic from the rest of our experiments. Hindi was excluded from our experiments with Llama2Chat because limited context length of 4k token.

\subsection{Metrics}
\label{subsec:metrics}
DIT assesses three separate and developmentally ordered moral schemas \cite{rest1999postconventional}. These schemas are identified as the Personal Interests schema, which combines elements from Kohlberg's Stages 2 and 3; the Maintaining Norms schema, derived from Kohlberg's Stage 4; and the Post-conventional schema, which draws from Kohlberg's Stages 5 and 6. The Post-conventional schema is equivalent to the original summary index known as the P-score.

The \textit{Personal Interest schema score} reflects an individual's tendency to make moral judgments based on their personal interests, desires, or self-benefit. A higher score in this context suggests that a person is more inclined to prioritize their own interests when making moral decisions. \textit{Maintaining norms score} measures a person's commitment to upholding societal norms and rules in their moral judgments. A higher score in this category indicates a greater emphasis on adhering to established norms and societal expectations when making ethical decisions. \textit{Post-conventionality score}/$p_{score}$ gauges a person's level of moral development, reflecting their inclination to make moral judgments based on advanced moral principles and ethical reasoning. A higher score in this category signifies a commitment to abstract ethical principles, justice, individual rights, and ethical values, transcending conventional societal norms.

In summary, the \textit{Personal Interest schema score} reflects self-centered moral reasoning, the \textit{Maintaining norms score} signifies a commitment to adhering to societal norms, and the \textit{Post-conventionality score} represents advanced moral reasoning based on ethical principles and values. Individual stage-wise score are defined as follows:

\begin{equation}
score_\theta = 10 \cdot \sum_{i=1}^{4} ((5-i)\cdot S_{i, \theta})
\end{equation}

where $S_{i, \theta}$ is defined as:

\[
S_{i, \theta} = 
\begin{cases}
    1 & \text{if the $i^{th}$ ranked statement $\in$ Stage-$\theta$} \\
    0 & \text{otherwise}
\end{cases}
\]

Therefore, \textit{Personal Interest schema score} =  $score_2 + score_3$, \textit{Maintaining norms score} = $score_4$ and $p_{score} = score_5 + score_6$. We present a working example of the score calculation in the Appendix Section~\ref{A:score}.

\section{Results and Observation}

\begin{figure*}[h] 
    \centering 
    \begin{subfigure}{0.3\linewidth} 
        \includegraphics[width=\linewidth]{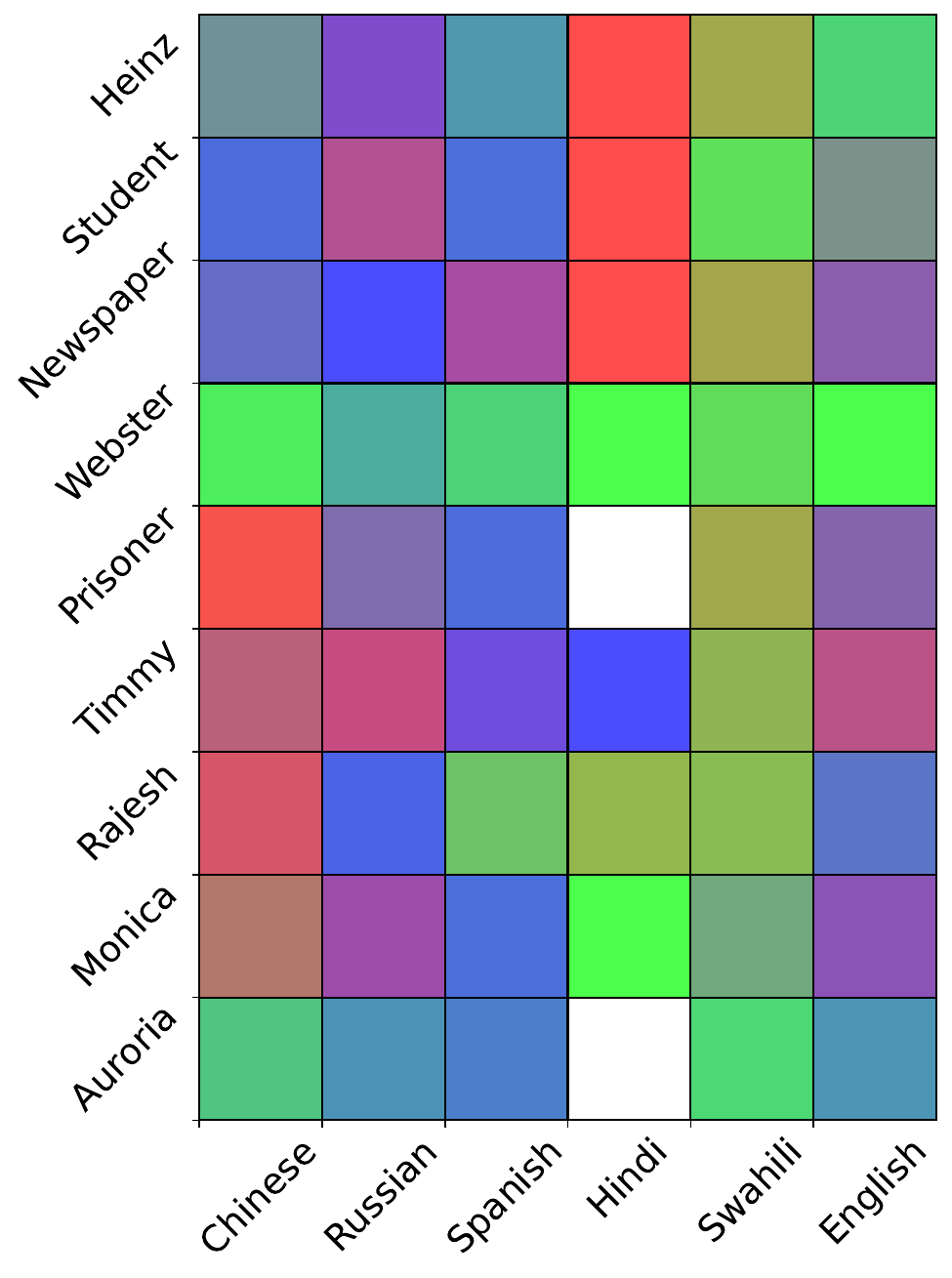} 
        \caption{ChatGPT} 
        \label{fig:chatgpt-resolution} 
    \end{subfigure}
    \hfill 
    \begin{subfigure}{0.3\linewidth}
    \includegraphics[height=2.55in, width=\linewidth]{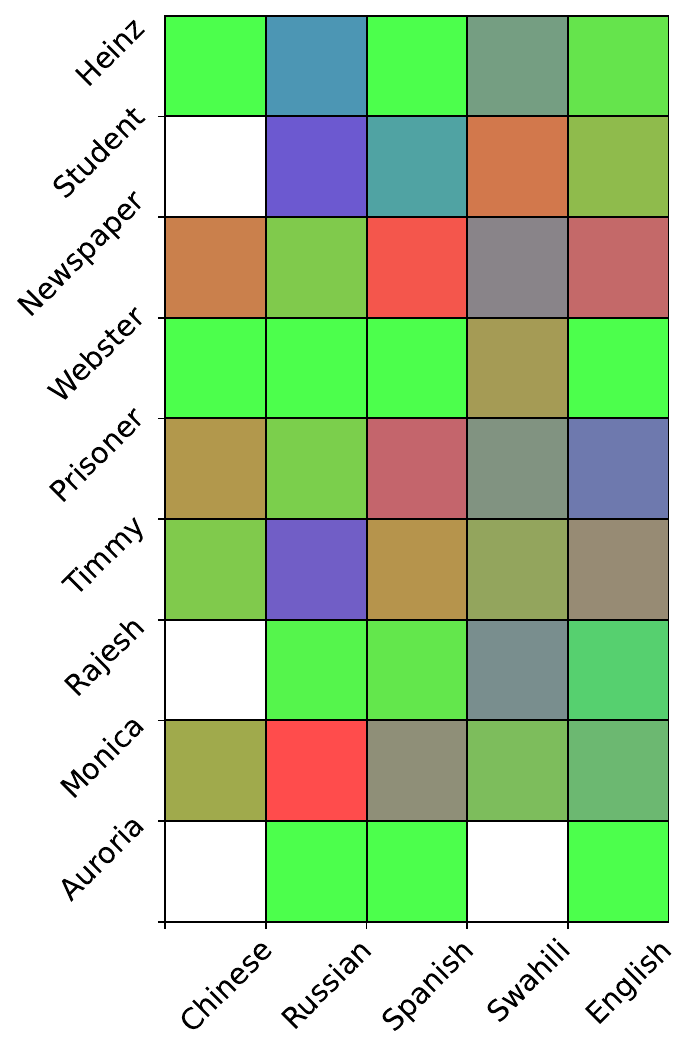}
    \caption{Llama2Chat-70B}
    \label{fig:llama-resolution}
    \end{subfigure}
    \hfill
    \begin{subfigure}{0.3\linewidth}
    \includegraphics[width=\linewidth]{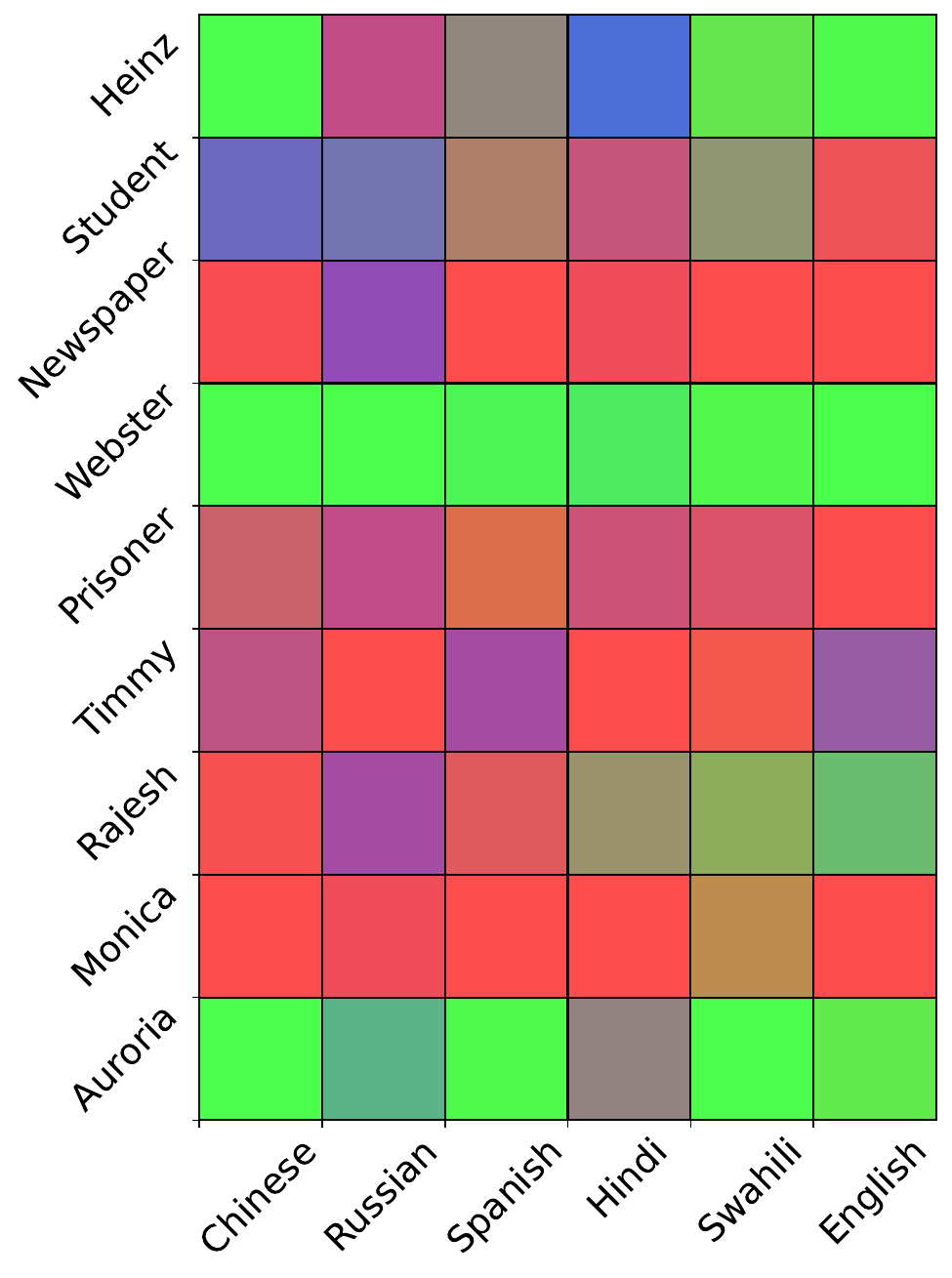}
    \caption{GPT-4}
    \label{fig:gpt4-resolution}
    \end{subfigure}
    \caption{Dilemma-specific resolution heatmaps across various languages for ChatGPT, Llama2chat-70B, and GPT-4. O1 is indicated in green, O2 in blue, and O3 in red. The heatmaps illustrate the number of instances where the models provided answers corresponding to O1, O2, or O3 for each language and dilemma based on the RGB component. White areas represent scenarios where no observations yielded an extractable resolution to the dilemma.} 
    \label{fig:resolution} 
\end{figure*}
\begin{figure*}[h] 
    \centering 
    \begin{subfigure}{0.293\linewidth} 
        \includegraphics[width=\linewidth]{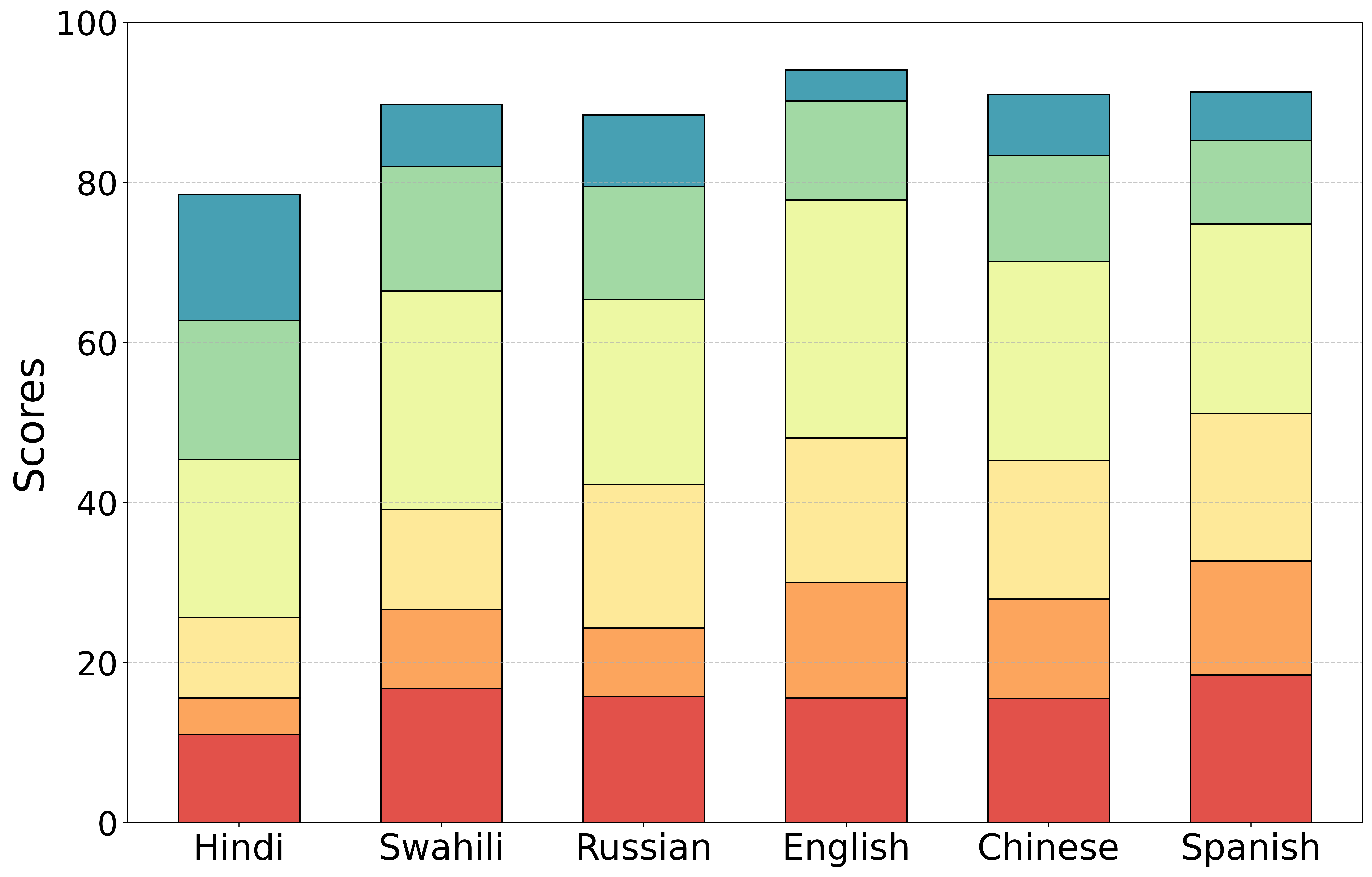} 
        \caption{ChatGPT} 
        \label{fig:chatgpt-stack}
    \end{subfigure}
    \hfill 
    \begin{subfigure}{0.293\linewidth}
    \includegraphics[width=\linewidth]{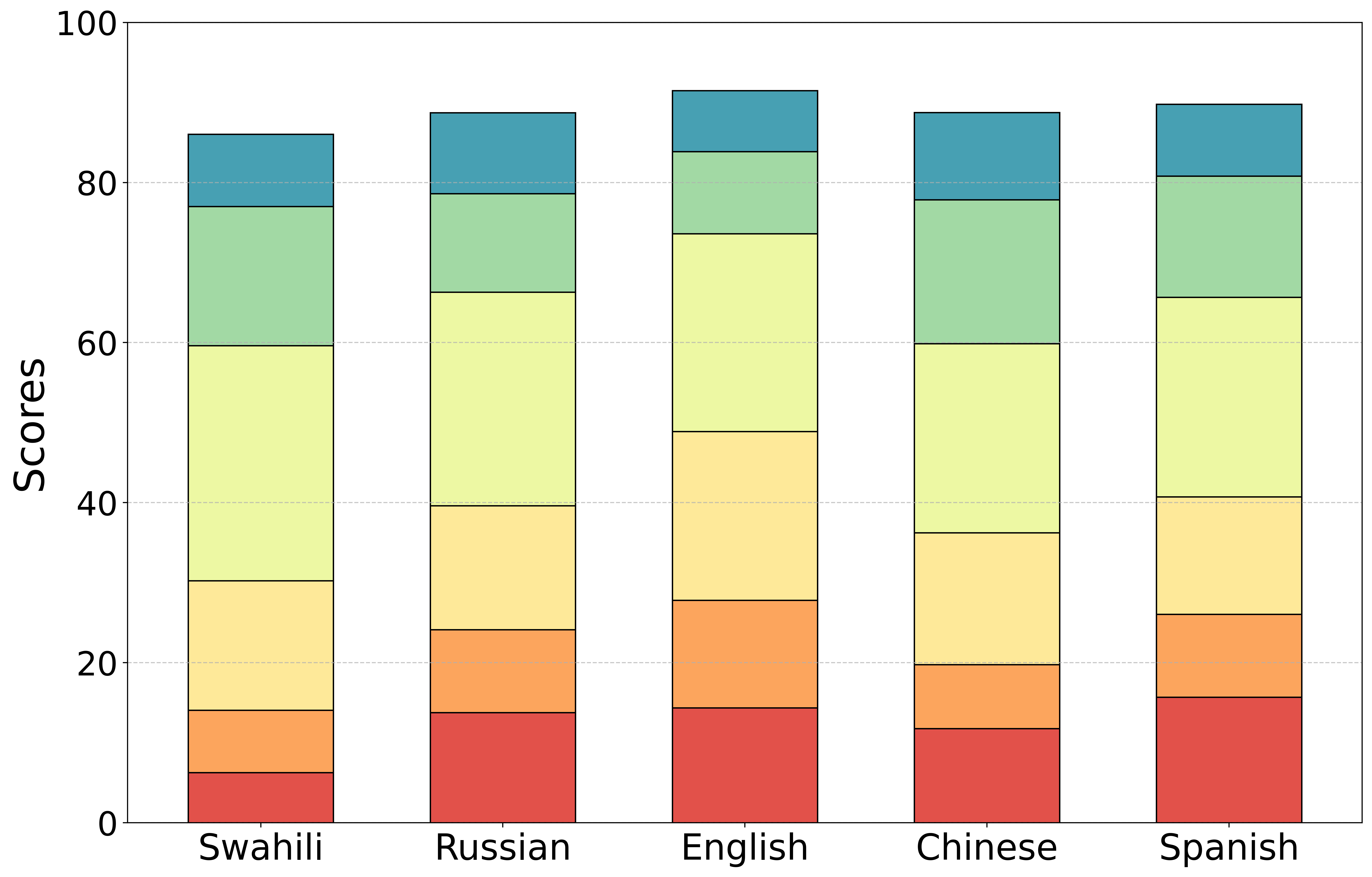}
    \caption{Llama2Chat-70B}
    \label{fig:llama-stack}
    \end{subfigure}
    \hfill
    \begin{subfigure}{0.378\linewidth}
    \includegraphics[width=\linewidth]{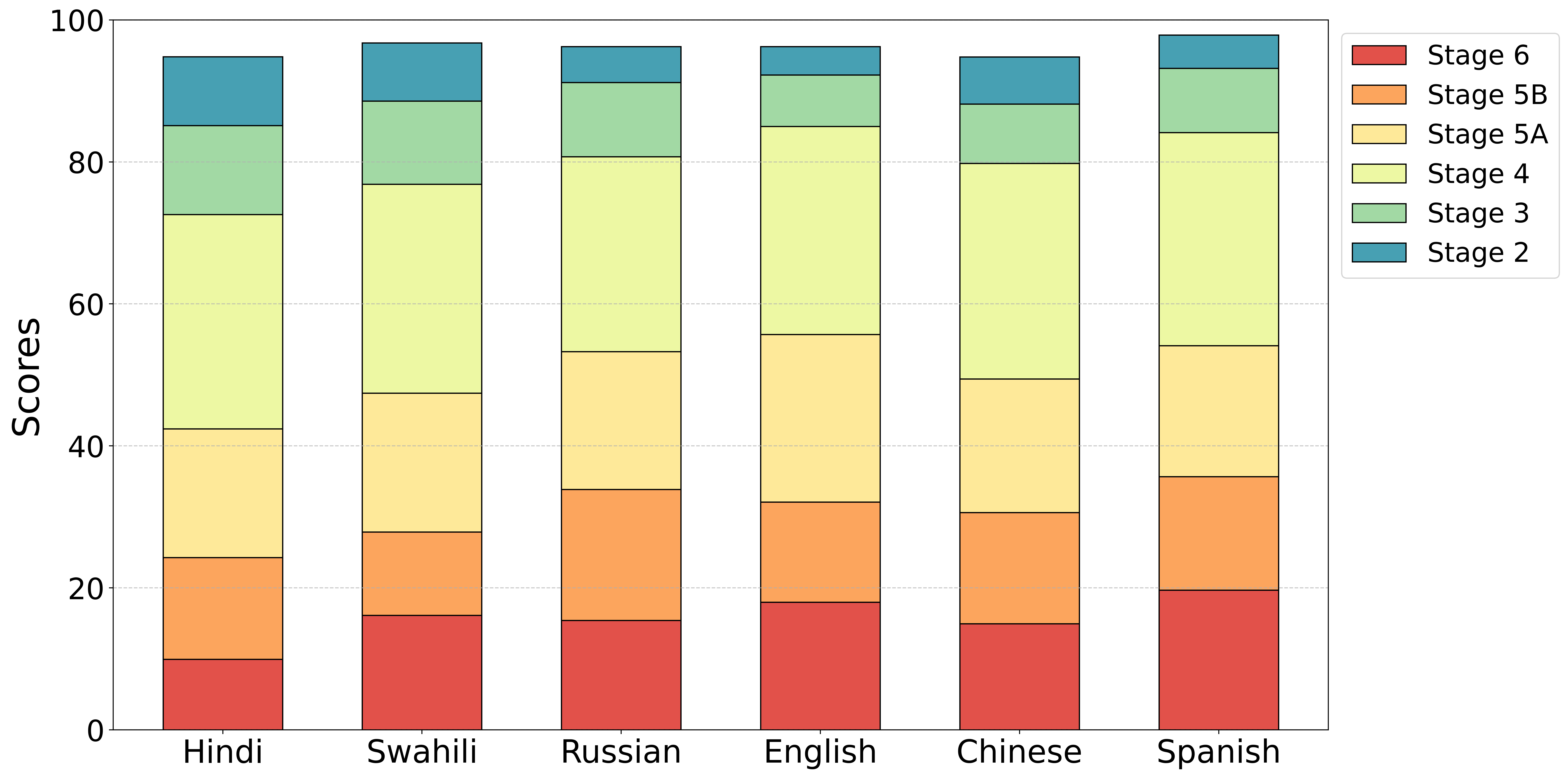}
    \caption{GPT-4}
    \label{fig:gpt4-stack}
    \end{subfigure}
    \caption{Overview of stage-wise scores for ChatGPT, Llama2Chat, and GPT-4, averaged across all moral dilemmas. The cumulative scores of the initial three tiers (Red, Orange, and Deep Yellow) is the $p_{score}$ or post-conventional morality score. The 4th tier (light yellow) signifies the Maintaining Norms schema score and the 5th and 6th tiers (green and blue) combined gives the Personal Interests schema score.}
    \label{fig:stackplots}
\end{figure*}

\subsection{Moral Judgment by the LLMs}
As discussed earlier, we consider {\em moral judgment} as the resolution of the dilemmas offered by the LLMs.  
Figure~\ref{fig:resolution} summarizes the resolutions from ChatGPT, Llama2chat-70B, and GPT-4 for different languages as a 2-dimensional color map across languages and dilemmas. Green color signifies the affirmative side of the resolution, as seen in \textit{``should share," ``should take," and ``should report"}, etc. Blue color represents the model's inability to adopt a specific moral stance, leading it to select the "Can't decide" option. Red color designates the negative side, exemplified by \textit{``should not share," ``should not take," and ``should not report"}, etc. 
Since the affirmative and negative statements are rather arbitrary, we list the actual values of these stances for each dilemma.  
\textit{
\begin{itemize}
\setlength\itemsep{0em} 
\item \textbf{Heinz} Should Heinz steal a drug from
an inventor in town to save his wife who is dying and needs the drug?
\item \textbf{Student} Should students take over an administration
building in protest of the Vietnam war?
\item \textbf{Newspaper} Should a student newspaper be stopped by a Principal of a high school when the newspaper stirs controversy in the community?
\item \textbf{Webster} Should a minority member be hired for a job when the community is biased?
\item \textbf{Prisoner} Should a man who escaped from prison but has since been leading an exemplary life be reported to authorities?
\item \textbf{Timmy} Should Timmy attend
his friend’s wedding instead of fixing an urgent bug that could put customers’ privacy at risk?
\item \textbf{Rajesh} Should Rajesh rent a house by hiding the secret of his non-vegetarian consumption at home from the vegetarian neighborhood?
\item \textbf{Monica} Should Monica give the first authorship
to Aisha despite having the major contribution?
\item \textbf{Auroria} Should the
country Auroria share its innovations and resources to it’s poor neighbor or profit off it’s huge investments in research?
\end{itemize}
}
\begin{figure*}[h] 
    \centering 
    \begin{subfigure}{0.27\linewidth} 
        \includegraphics[width=\linewidth]{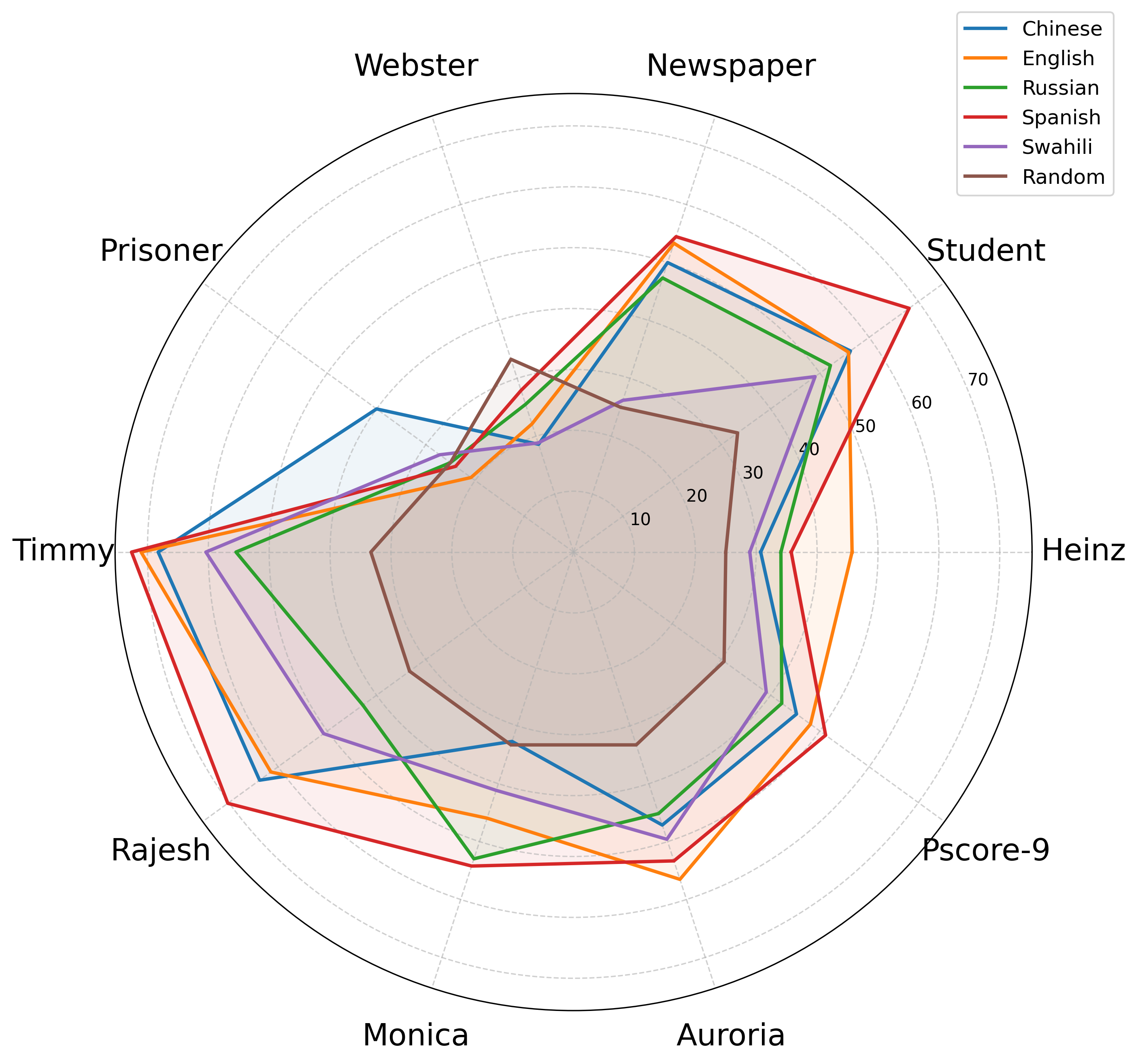} 
        \caption{P-Scores for ChatGPT} 
        \label{fig:chatgpt-radar} 
    \end{subfigure}
    \hfill 
    \begin{subfigure}{0.27\linewidth}
    \includegraphics[width=\linewidth]{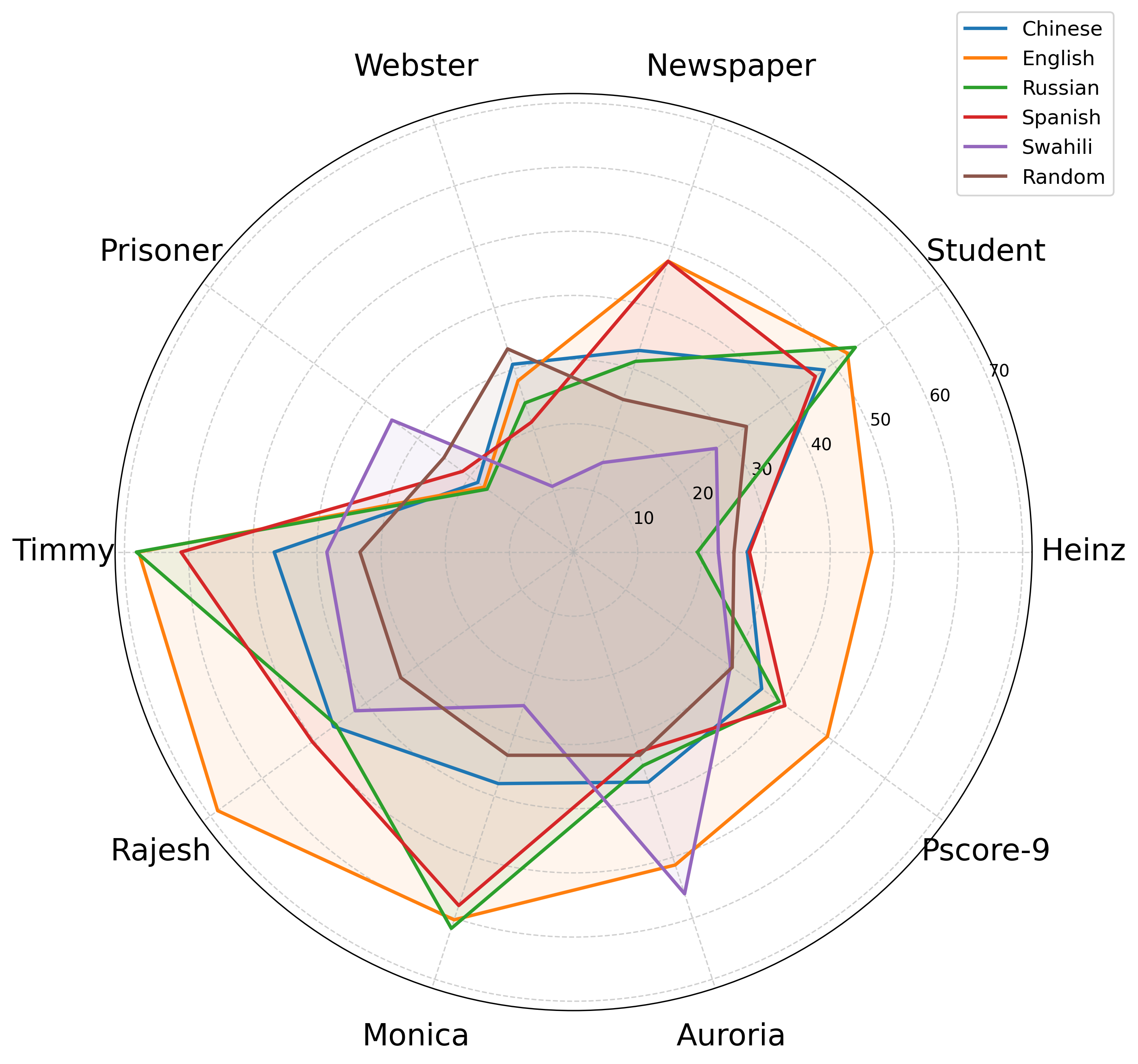}
    \caption{P-Scores for Llama2Chat-70B}
    \label{fig:llama-radar}
    \end{subfigure}
    \hfill
    \begin{subfigure}{0.27\linewidth}
    \includegraphics[width=\linewidth]{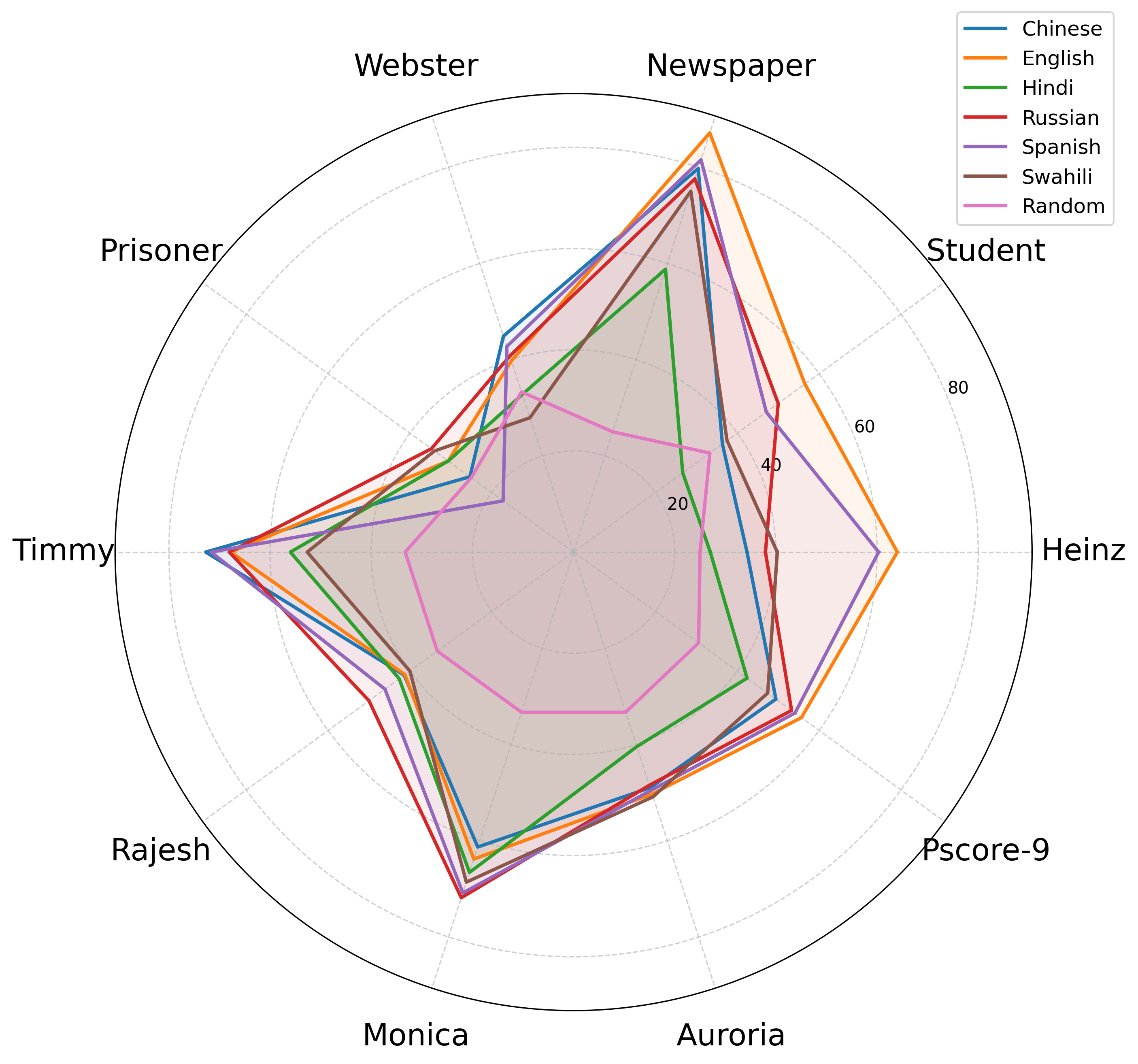}
    \caption{P-Scores for GPT-4}
    \label{fig:gpt4-radar}
    \end{subfigure}
    
    \vspace{5pt}
    \begin{subfigure}{0.27\linewidth} 
        \includegraphics[width=\linewidth]{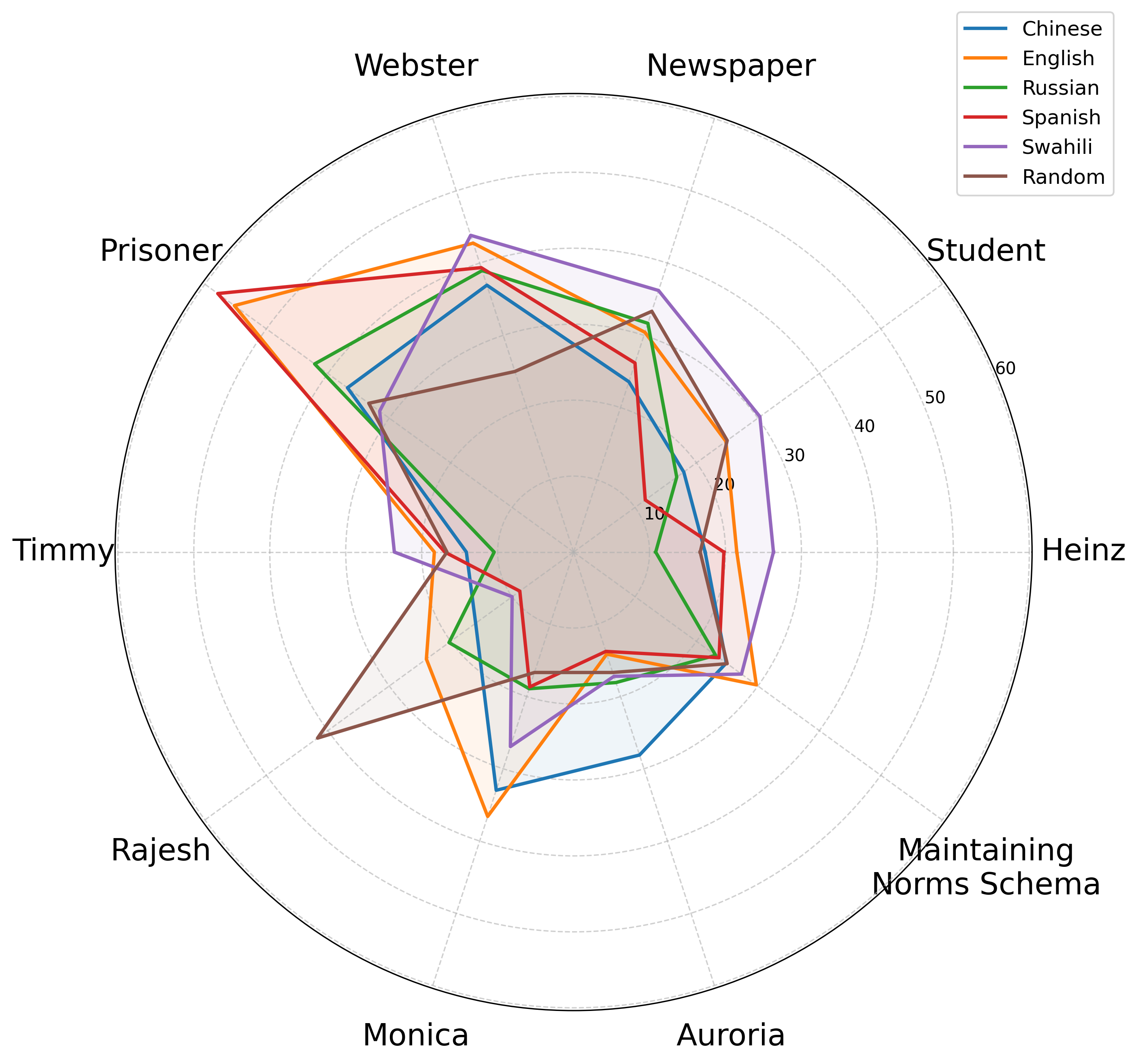} 
        \caption{Maintaining Norms Schema score for ChatGPT} 
        \label{fig:chatgpt-radar} 
    \end{subfigure}
    \hfill 
    \begin{subfigure}{0.27\linewidth}
    \includegraphics[width=\linewidth]{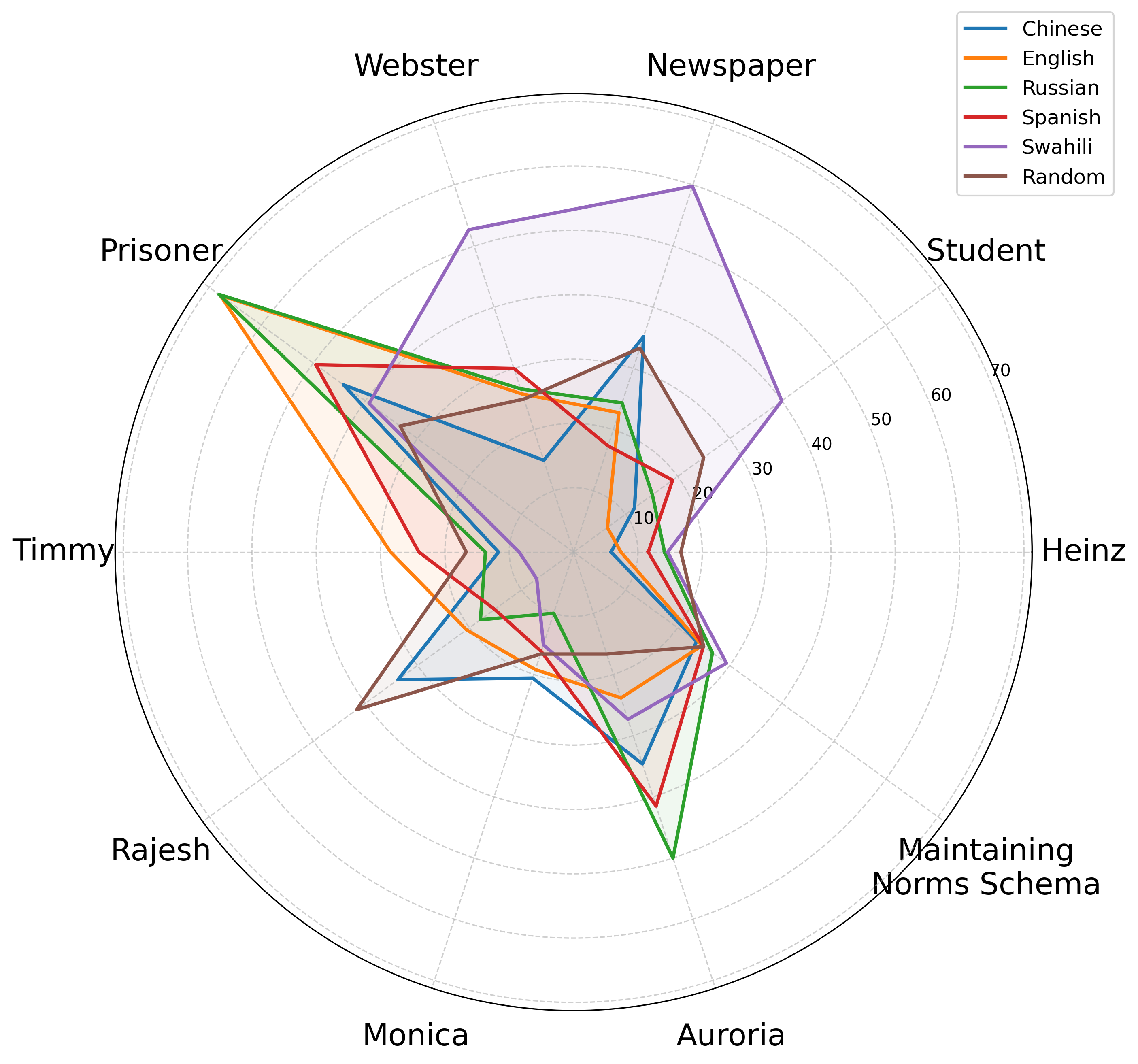}
    \caption{Maintaining Norms Schema score for Llama2Chat-70B}
    \label{fig:llama-radar}
    \end{subfigure}
    \hfill
    \begin{subfigure}{0.27\linewidth}
    \includegraphics[width=\linewidth]{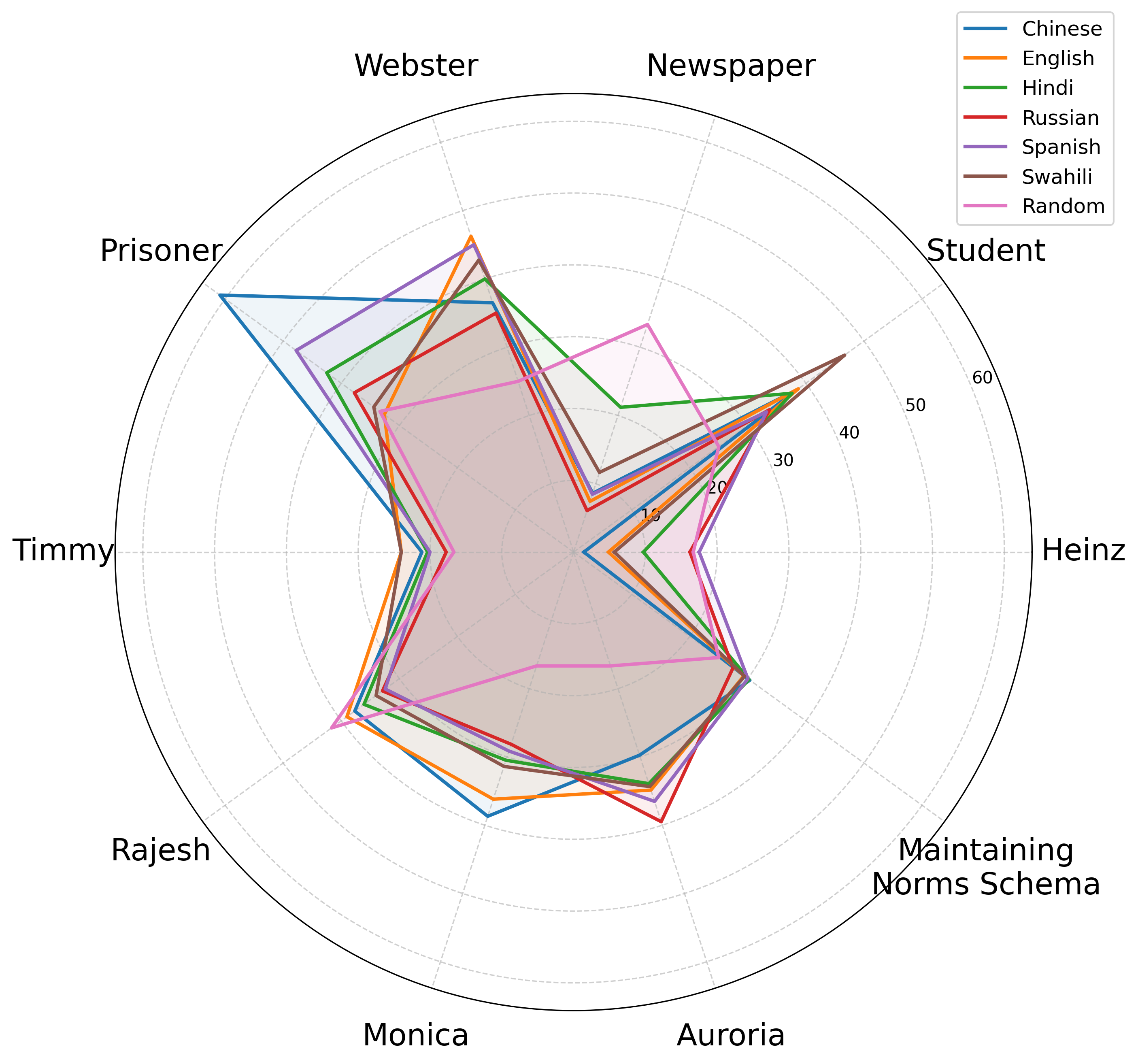}
    \caption{Maintaining Norms Schema Score for GPT-4}
    \label{fig:gpt4-radar}
    \end{subfigure}
    
    \vspace{5pt}
    \begin{subfigure}{0.27\linewidth} 
        \includegraphics[width=\linewidth]{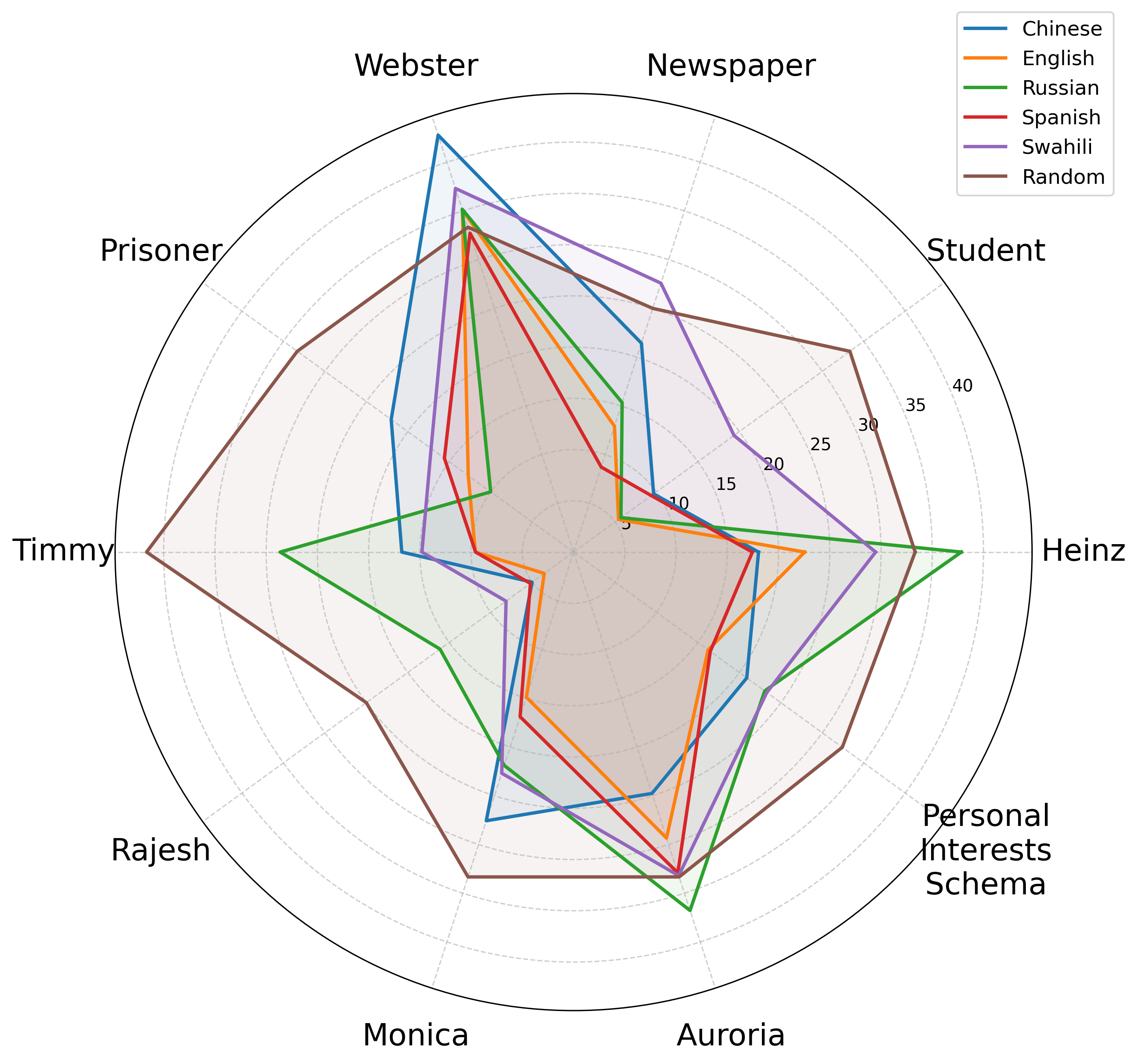} 
        \caption{Personal Interests Schema score for ChatGPT} 
        \label{fig:chatgpt-radar} 
    \end{subfigure}
    \hfill 
    \begin{subfigure}{0.27\linewidth}
    \includegraphics[width=\linewidth]{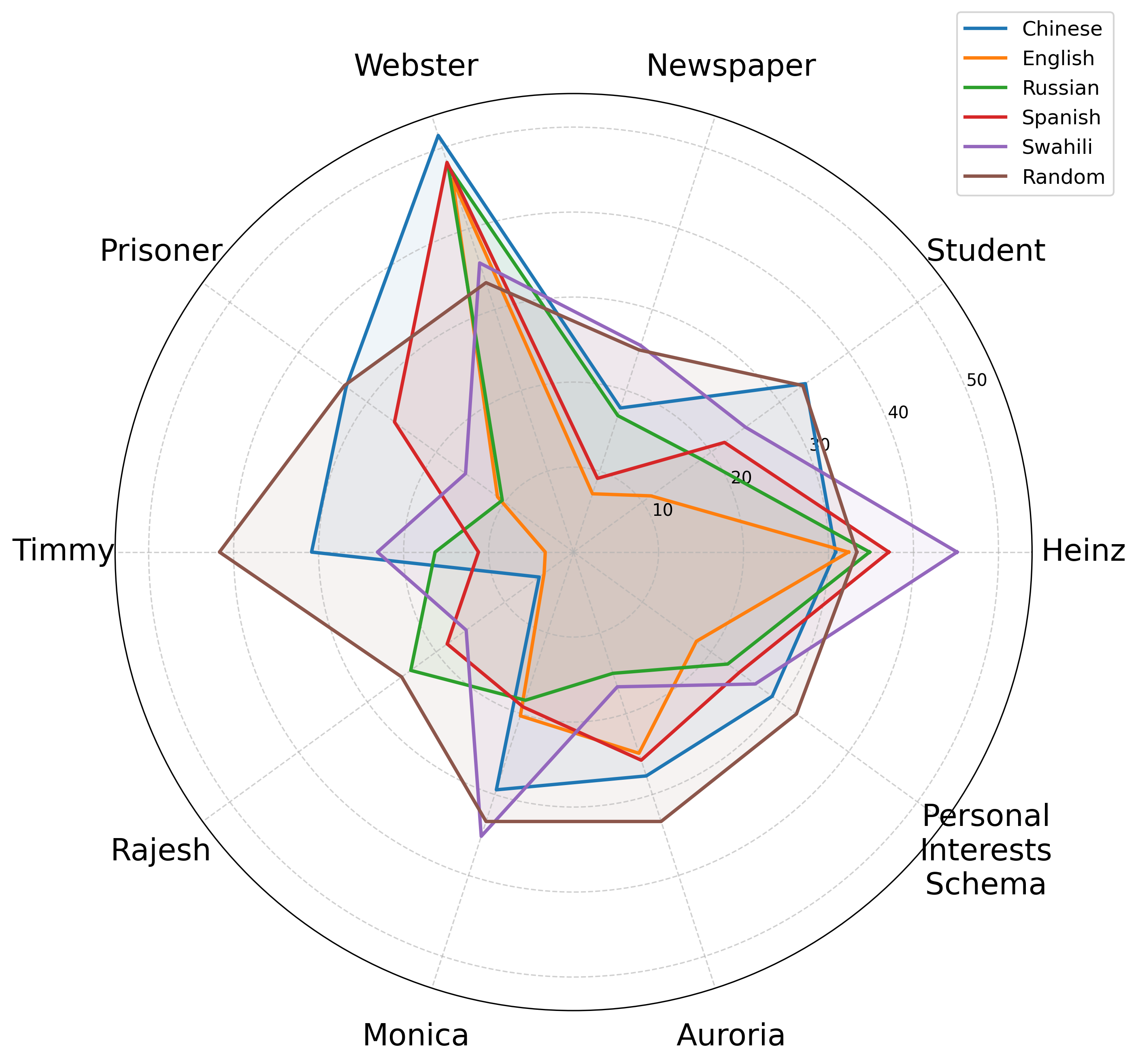}
    \caption{Personal Interests Schema score for Llama2Chat-70B}
    \label{fig:llama-radar}
    \end{subfigure}
    \hfill
    \begin{subfigure}{0.27\linewidth}
    \includegraphics[width=\linewidth]{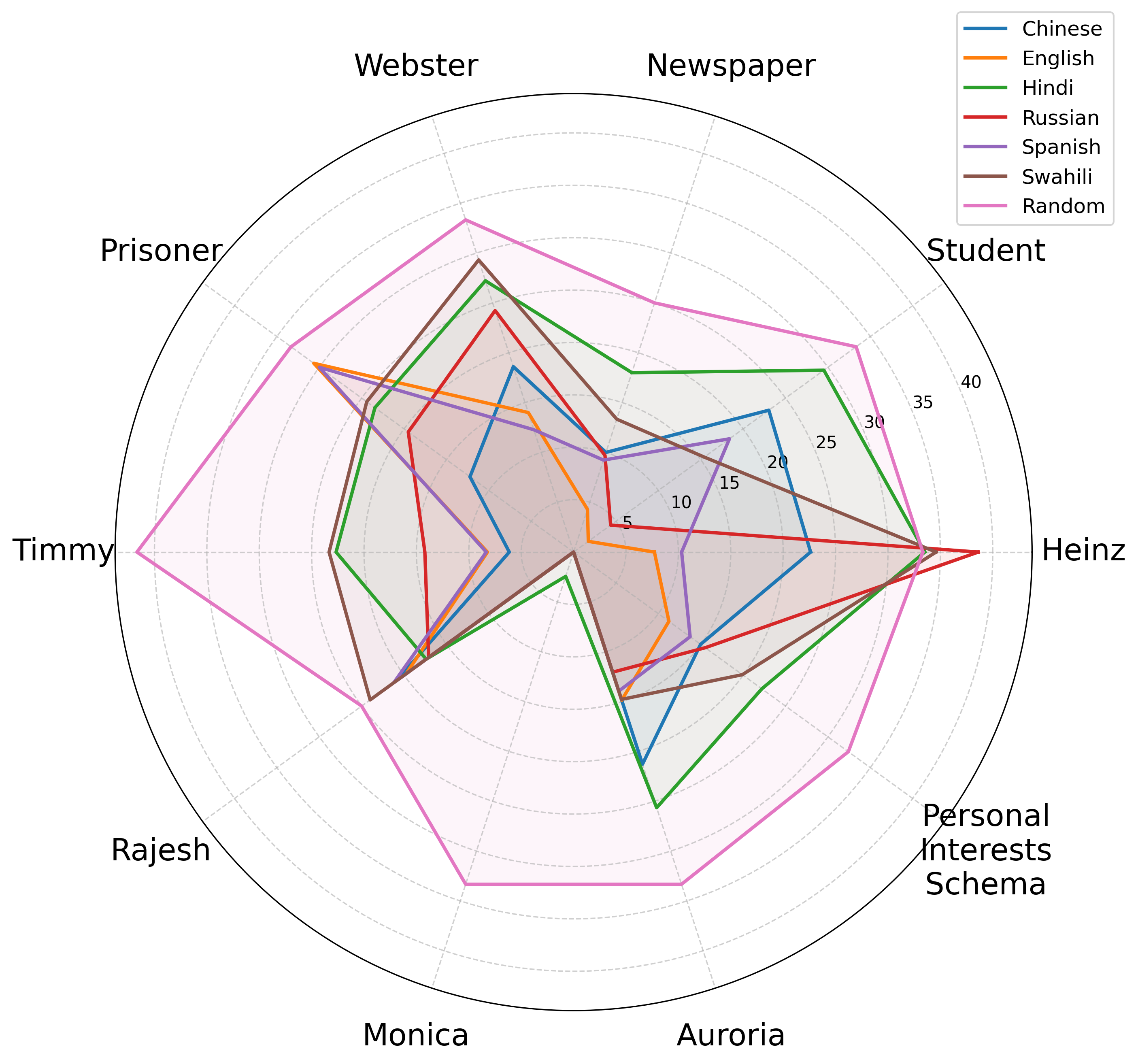}
    \caption{Personal Interests Schema score for GPT-4}
    \label{fig:gpt4-radar}
    \end{subfigure}

    \caption{Comparing dilemma-specific and overall P-scores among ChatGPT, Llama2Chat, and GPT-4, versus the random baselines, across five languages for ChatGPT and Llama2Chat (excluding Hindi) and six languages for GPT-4.} 
    \label{fig:radarplots} 
\end{figure*}
It is evident from the Figure~\ref{fig:resolution} that GPT-4 exhibits a significantly higher level of consensus in the resolutions across different languages, in comparison to Llama2Chat and ChatGPT. Quite intriguingly, GPT-4 predominantly yields "O3" responses, whereas Llama2Chat tends to produce more "O1" responses, and ChatGPT more O2 (``cant' decide") responses especially for high-resource languages like English, Chinese, Russian, and Spanish. It's worth noting that all models and languages converge towards an O1 response for the Webster and Auroria dilemmas. In contrast, for the Student dilemma we observe a considerable degree of variation in the resolutions across languages for all models.

Comparing the resolution patterns across languages, we observe that for all models, resolution in English and Spanish are similar to each other. For Llama2Chat and GPT-4, moral judgments in Spanish and Chinese are similar, while those in Russian and English are most different. In contrast, for ChatGPT, Russian and English resolutions are quite similar, while resolutions in Swahili and Russian, and in Swahili and Chinese are most dissimilar. Overall, moral judgments in Russian seem to disagree most with that in other languages, especially for GPT-4 and Llama2Chat.

It is interesting to speculate the potential reasons behind these differences. It is possible that for low-resource languages like Hindi and Swahili, the model does not have exposure to enough pre-training and fine-tuning data to learn the typical cultural values for the L1 speakers of these languages; neither the LLMs are capable of performing complex reasoning and processing in these languages, as has been shown by several recent multilingual benchmarking studies~\cite{ahuja2023mega, wang2023seaeval}. Therefore, for these languages, the resolutions are either random or a direct translation of the moral resolutions in a high resource language such as English (as if English was the L1 of the LLM, and languages for which it had very limited proficiency, such as third language - L3 or fourth language - L4, it translated the input to English, reasoned over the translated input and translated the response back to the Language). Indeed, Llama2Chat responded in English for Swahili and even for Chinese. 

\begin{figure}
    \centering
    \includegraphics[scale=0.33, trim=15 3 20 10,clip]{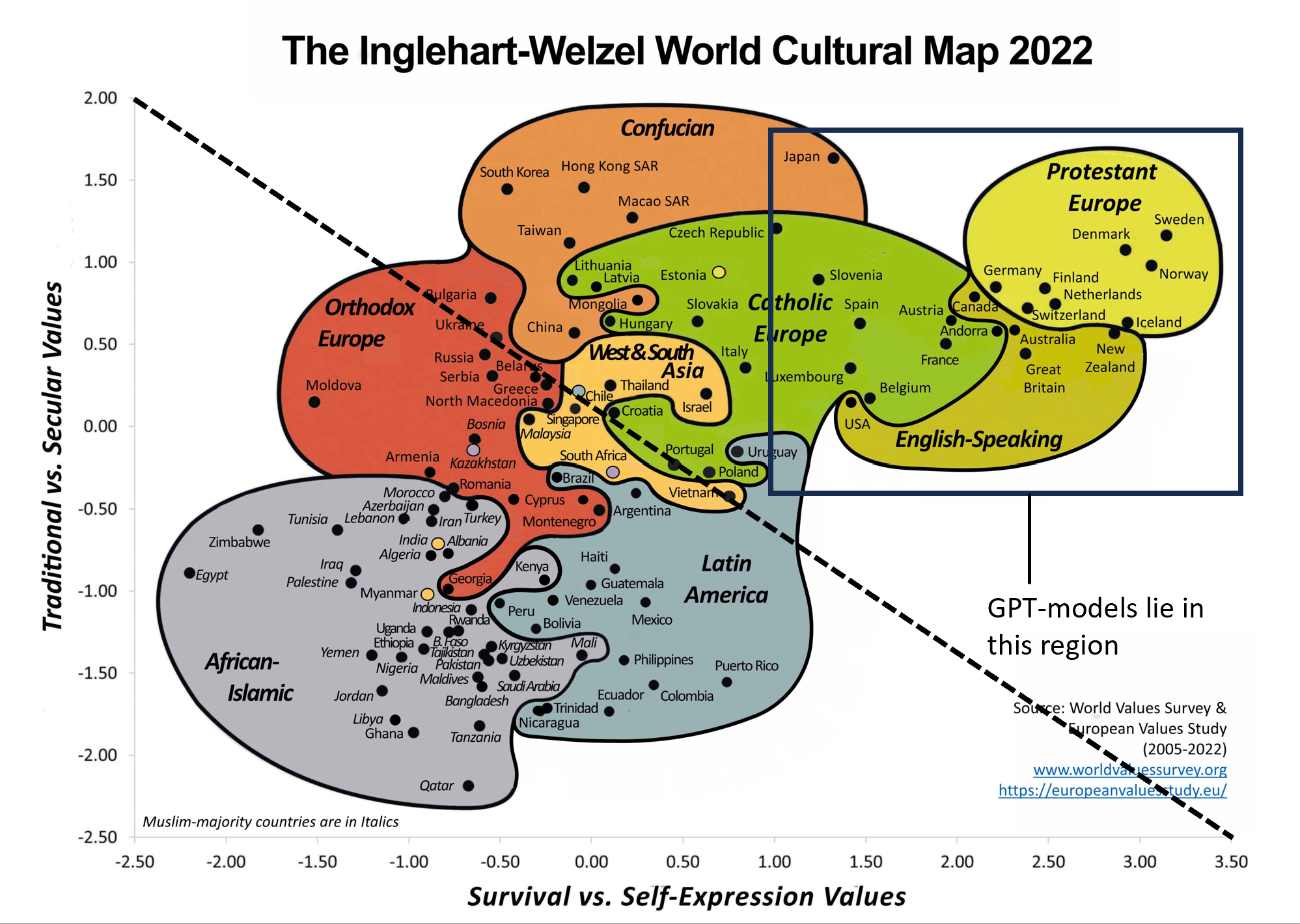}
    \caption{An illustration of contemporary Language Models with the world cultural map \cite{rao2023ethical}.}
    \label{fig:worldview}   
\end{figure}

On the other hand, for a relatively high resource language, like Spanish, Chinese and Russian, the LLMs might have had sufficient exposure to data from which it could learn the cultural values of the L1 speakers of these languages. According to the World Value Survey, Russia (orthodox European) is farthest from English speaking countries on the value map (see Fig~\ref{fig:worldview}), and thus, perhaps, elicits the most dissimilar moral judgments compared to English. On the other hand, Spain (Catholic Europe) is closest (among the languages we studied) to English on the value map, followed by Chinese and thus, these languages elicit similar responses to that of English. 

Interestingly, the resolutions in Russian and Chinese significantly differ from each other for all models, despite Russia and China being closely placed on the value map. A possible explanation for this could be as follows. As \citet{rao2023ethical} speculate, the LLMs seem to align to the values on the right-upper triangle of the map (above the dashed diagonal line in Fig~\ref{fig:worldview}). China, Spain and English speaking countries are on the upper-right triangle, while Russia falls into the lower-left triangle, which might explain the differences in the moral judgments. In other words, the behavior of the LLMs seem to change for languages on the two sides of the dashed line, which could also be an artifact of the nature of these specific dilemmas.

\subsection{Moral Reasoning by LLMs}

As discussed in Section~\ref{subsec:moralreasoning}, moral reasoning is how people think through what's right or wrong by using their values and ethical principles. It involves critical thinking and understanding different ethical ideas, using both logical and emotional thoughts to make ethical choices \cite{richardson2003moral}. In simpler terms, it's the process behind forming moral judgments. \citet{Rest1986B} shows how moral reasoning can be understood with the help of DIT scores from a rationalist perspective.

In Figure~\ref{fig:stackplots}, we can see the stages of cognitive moral development for these models for different languages. Across all models, CMD tends to be concentrated in the post-conventional morality stage, with an exception of ChatGPT for Hindi where its moral reasoning is predominantly centered around the {\em personal interests} schema and Llama2Chat for Swahili, where it is concentrated around the {\em maintaining norms} schema score. For both ChatGPT and Llama2Chat, there is a more balanced distribution between the two moral schemas, {\em maintaining norms} and {\em personal interest}.
The average (over all languages) maintaining norms schema scores of Llama2Chat and ChatGPT are 25.68 and 22.17 respectively, while the average personal interest schema scores are 23.93 and 24.74 respectively. 
GPT-4 exhibits a notably different pattern. Its values for these schemas are significantly lower compared to the average post-conventional schema score (or P-score). For GPT-4. Thus, compared to ChatGPT and Llama2Chat, GPT-4 has a more developed moral reasoning capability for all the languages studied. The lowest P-score was observed for Hindi, which too is greater than 40, and is in the range of P-scores observed in adult humans \cite{Rest1986B}. 

Figure~\ref{fig:radarplots} shows the P-scores, maintaining norms schema scores and personal interest schema scores for all languages across all dilemmas and models. We also mark the random baseline score (when the top 4 statements are picked at random from the 12 moral considerations by a model) for each of these schemas.
We note that for Webster dilemma all models had consensus in moral judgment, however the moral reasoning for resolving this dilemma lies in the personal interests schema, indicating rather underdeveloped moral reasoning. Interestingly, for Heinz dilemma, GPT-4 and ChatGPT exhibit high score in the personal interest schema for all languages, but Llama2Chat shows high variation across languages.
We further note that the all the models take the maintaining social norms perspective (Stage 4 specific) while resolving the Prisoner dilemma with a slight variation across language. In short, even though, on average we observe post-conventional or near post-conventional moral reasoning abilities in GPT-4 for all languages, and near post-conventional moral reasoning for all languages except Swahili for Llama2Chat, for certain dilemmas the models display conventional or pre-conventional morality. 

Due to paucity of space, we omit several other results. Table~\ref{t:differencedilemma} in the Appendix presents a comprehensive report of the P-scores (the most common single index used in DIT based studies) of the LLMs across all dilemmas and languages. We also conducted Mann-Whitney U Tests of statistical significance over various runs. Wherever the P-scores in English are statistically significantly different ($p<0.05$) from that in another language, the numbers are shown in bold. The salient observations from this analysis are: (a) For Webster and Prisoner dilemma, there is no significant difference in P-scores of the models across languages; (b) GPT-4's P-scores across languages for Rajesh and Auroria dilemmas show no significant differences; and (c) for all models, we observe the maximum statistically significant difference in P-scores across languages for the Heinz dilemma, followed by the Newspaper dilemma.

\section{Discussion and Conclusion}
In this first of its kind study of multilingual moral reasoning assessment of LLMs, we observe that quite unsurprisingly, the moral reasoning capability, as quantified by the DIT stage scores, of LLMs is highest for English, followed by Spanish, Russian and Chinese, and lowest for Hindi and Swahili. GPT-4 emerges as the most capable multilingual moral reasoning model with less pronounced differences in its capabilities in different languages. Nevertheless, we also observe remarkable variation in moral judgments and reasoning abilities across dilemmas. 

Our work opens up several intriguing questions about LLMs moral reasoning, and the role of language and cultural values that were presented in form of textual data during the pre-training, instruction fine-tuning and RLHF stages of the model. Since these datasets are often unavailable for scrutiny (especially true for ChatGPT and GPT-4), we can only speculate the reasons for the differences. It will be interesting to design specific experiments to probe further into the hypotheses and postulates that have been offered as plausible explanations in this paper. 

\section*{Limitations}
This study has some notable limitations. Firstly, the evaluation framework we used from this work \cite{tanmay2023probing} may contain bias, as it include some dilemmas specifically designed from a Western perspective. Although other dilemmas also consider diverse cultural viewpoints, the complexity of ethical perspectives across cultures may not be fully captured. Secondly, our study's scope is limited to a few languages, primarily focusing on linguistic diversity, which may restrict the generalizability of our findings to languages not included. Additionally, the use of Google Translator for multilingual dilemma translation carries the potential for translation errors. Despite these limitations, our research offers insights into cross-cultural ethical decision-making of LLMs in diverse languages, highlighting the need for future investigations to address these constraints and strengthen the robustness of our findings.

\section*{Ethical Concerns}
Our results show that GPT-4 is a post-conventional moral reasoner (with scores comparable to philosophers and graduate students) across most of the languages studied, and it is at least as good as an average adult human for all languages on moral reasoning tasks. This might lead people to think that GPT-4 or similar models can be used for making real life ethical decisions. However, this could be very dangerous as, firstly, our experimental setup is limited to only 9 dilemmas covering a small set of cultural contexts and values; secondly, our experiments are limited to 6 languages, which cannot and should not be generalized to the model's performance to other languages beyond those tested. We believe that the current work does not provide sufficient and reliable ground for using LLMs for making moral judgments.

\bibliography{custom}
\newpage
\newpage
\appendix

\section{Appendix}
\label{sec:appendix}

\subsection{Score-Calculation}
\label{A:score}
The DIT presents a moral dilemma and 12 statements that correspond to different stages of Kohlberg’s Cognitive Moral Development. The respondent has to answer three questions. Question 1 requires the respondent to give the moral judgment for the dilemma. Question 2 asks the respondent to evaluate the significance of each statement in making the moral judgment. The respondent does not know which statement belongs to which stage of CMD. Question 3 asks the test taker to choose the 4 most important statements (ranked in order of importance) that influenced the moral judgment.

There are three metrics in the DIT: Post-conventional schema (Stage 5 and 6), Maintaining Norms schema (Stage 4), and Personal Interests schema (Stage 2 and 3). The test taker has to choose the 4 most important statements out of the 12, and rank them in order of importance. For example, suppose the respondent chooses statement \#11 as the most important statement, \#7 as the second most important, \#9 as third most important and \#2 as the fourth most important statement.

The statements belong to different stages of CMD, but the respondent does not know which stage each statement belongs to. Consider the case where the stages for a particular dilemma are as follows: \#2 (Stage 3), \#7 (Stage 4), \#9 (Stage 6) and \#11 (Stage 5). The scores for each stage are calculated as follows:

\noindent
Stage 6 score: 

$10\times(4\times0+3\times0+2\times1+1\times0)=20$

\noindent
Stage 5 score: 

$10\times(4\times1+3\times0+2\times0+1\times0)=40$

\noindent
Stage 4 score: 

$10\times(4\times0+3\times1+2\times0+1\times0)=30$

\noindent
Stage 3 score: 

$10\times(4\times0+3\times0+2\times0+1\times1)=10$

\noindent
Stage 2 score: 

$0$

The final scores for each scores are as follows:

\noindent
Personal Interests schema score:

$score_2 + score_3 = 0 + 10 = 10$

\noindent
Maintaining Norms schema score:

$score_4 = 30$

\noindent
Post-conventional schema score:

$P_{score} = score_5 + score_6 = 40 + 20 = 60$

\begin{table*}
\fontsize{8.9}{11pt}\selectfont
\centering
\setlength{\tabcolsep}{2pt}
\begin{tabular}{l l l l l l l l l l ll} 
\toprule
\textbf{Model} & \textbf{Lang}. & \textbf{Heinz} & \textbf{Student} & \textbf{Newspaper} & \textbf{Webster} & \textbf{Prisoner} & \textbf{Timmy} & \textbf{Rajesh} & \textbf{Monica} & \textbf{Auroria}  &\textbf{P-Score}\\
\midrule
  \multirow{6}{*}{\rotatebox[origin=r]{90}{\textbf{ChatGPT}}}& en & 45.74& 55.83& 53.33& 22.13& 20.83& 71.04& 61.46& 45.96&56.52 &48.09\\
& zh & \textbf{30.73}\textcolor{Maroon}{\textsubscript{$\downarrow$32.8}}& 56.21\textcolor{DarkGreen}{\textsubscript{$\uparrow$0.7}}& 50.00\textcolor{Maroon}{\textsubscript{$\downarrow$6.3}}& 18.61\textcolor{Maroon}{\textsubscript{$\downarrow$15.9}}& \textbf{40.00}\textcolor{DarkGreen}{\textsubscript{$\uparrow$92.0}}& 68.24\textcolor{Maroon}{\textsubscript{$\downarrow$4.0}}& 63.75\textcolor{DarkGreen}{\textsubscript{$\uparrow$3.7}}& \textbf{32.70}\textcolor{Maroon}{\textsubscript{$\downarrow$28.8}}& \textbf{47.14}\textcolor{Maroon}{\textsubscript{$\downarrow$16.6}} &45.26\textcolor{Maroon}{\textsubscript{$\downarrow$5.9}}\\
 & hi & \textbf{20.00}\textcolor{Maroon}{\textsubscript{$\downarrow$56.3}}& 44.00\textcolor{Maroon}{\textsubscript{$\downarrow$21.2}}& \textbf{10.00}\textcolor{Maroon}{\textsubscript{$\downarrow$81.3}}& 31.11\textcolor{DarkGreen}{\textsubscript{$\uparrow$40.6}}& -- & 40.00\textcolor{Maroon}{\textsubscript{$\downarrow$43.7}}& \textbf{20.00}\textcolor{Maroon}{\textsubscript{$\downarrow$67.5}}& 35.56\textcolor{Maroon}{\textsubscript{$\downarrow$22.6}}& \textbf{30.00}\textcolor{Maroon}{\textsubscript{$\downarrow$46.9}} &25.63\textcolor{Maroon}{\textsubscript{$\downarrow$46.7}}\\
 & ru & \textbf{34.05}\textcolor{Maroon}{\textsubscript{$\downarrow$25.6}}& 52.14\textcolor{Maroon}{\textsubscript{$\downarrow$6.6}}& 47.33\textcolor{Maroon}{\textsubscript{$\downarrow$11.3}}& 25.52\textcolor{DarkGreen}{\textsubscript{$\uparrow$15.3}}& 25.00\textcolor{DarkGreen}{\textsubscript{$\uparrow$20.0}}& \textbf{55.45}\textcolor{Maroon}{\textsubscript{$\downarrow$21.9}}& \textbf{42.78}\textcolor{Maroon}{\textsubscript{$\downarrow$30.4}}& 52.97\textcolor{DarkGreen}{\textsubscript{$\uparrow$15.3}}& \textbf{45.16}\textcolor{Maroon}{\textsubscript{$\downarrow$20.1}} &42.27\textcolor{Maroon}{\textsubscript{$\downarrow$12.1}}\\
 & es & \textbf{35.74}\textcolor{Maroon}{\textsubscript{$\downarrow$21.9}}& \textbf{68.12}\textcolor{DarkGreen}{\textsubscript{$\uparrow$22.0}}& 54.47\textcolor{DarkGreen}{\textsubscript{$\uparrow$2.1}}& 27.92\textcolor{DarkGreen}{\textsubscript{$\uparrow$26.2}}& 23.95\textcolor{DarkGreen}{\textsubscript{$\uparrow$15.0}}& 72.61\textcolor{DarkGreen}{\textsubscript{$\uparrow$2.2}}& \textbf{70.21}\textcolor{DarkGreen}{\textsubscript{$\uparrow$14.2}}& \textbf{54.22}\textcolor{DarkGreen}{\textsubscript{$\uparrow$18.0}}& 53.33\textcolor{Maroon}{\textsubscript{$\downarrow$5.6}} &51.18\textcolor{DarkGreen}{\textsubscript{$\uparrow$6.4}}\\
 & sw & \textbf{28.95}\textcolor{Maroon}{\textsubscript{$\downarrow$36.7}}& 49.03\textcolor{Maroon}{\textsubscript{$\downarrow$12.2}}& \textbf{26.21}\textcolor{Maroon}{\textsubscript{$\downarrow$50.9}}& 18.85\textcolor{Maroon}{\textsubscript{$\downarrow$14.8}}& 27.19\textcolor{DarkGreen}{\textsubscript{$\uparrow$30.5}}& \textbf{60.40}\textcolor{Maroon}{\textsubscript{$\downarrow$15.0}}& \textbf{50.74}\textcolor{Maroon}{\textsubscript{$\downarrow$17.4}}& 41.15\textcolor{Maroon}{\textsubscript{$\downarrow$10.5}}& 49.60\textcolor{Maroon}{\textsubscript{$\downarrow$12.3}} &39.12\textcolor{Maroon}{\textsubscript{$\downarrow$18.7}}\\
  \midrule
\multirow{5}{*}{\rotatebox[origin=r]{90}{\textbf{Llama2Chat}}}  & en& 46.47& 52.75& 47.67& 28.06& 17.23& 67.78& 68.57& 60.26&51.28 &48.9\\
& zh & \textbf{27.08}\textcolor{Maroon}{\textsubscript{$\downarrow$41.7}}& 48.29\textcolor{Maroon}{\textsubscript{$\downarrow$8.5}}& \textbf{33.04}\textcolor{Maroon}{\textsubscript{$\downarrow$30.7}}& 30.77\textcolor{DarkGreen}{\textsubscript{$\uparrow$9.7}}& 18.46\textcolor{DarkGreen}{\textsubscript{$\uparrow$7.1}}& \textbf{46.67}\textcolor{Maroon}{\textsubscript{$\downarrow$31.2}}& \textbf{46.25}\textcolor{Maroon}{\textsubscript{$\downarrow$32.6}}& \textbf{37.94}\textcolor{Maroon}{\textsubscript{$\downarrow$37.0}}& \textbf{37.69}\textcolor{Maroon}{\textsubscript{$\downarrow$26.5}} &36.24\textcolor{Maroon}{\textsubscript{$\downarrow$25.9}}\\
 & ru & \textbf{19.31}\textcolor{Maroon}{\textsubscript{$\downarrow$58.5}}& 54.29\textcolor{DarkGreen}{\textsubscript{$\uparrow$2.9}}& \textbf{31.25}\textcolor{Maroon}{\textsubscript{$\downarrow$34.5}}& 24.44\textcolor{Maroon}{\textsubscript{$\downarrow$12.9}}& 16.67\textcolor{Maroon}{\textsubscript{$\downarrow$3.3}}& 68.15\textcolor{DarkGreen}{\textsubscript{$\uparrow$0.6}}& \textbf{45.79}\textcolor{Maroon}{\textsubscript{$\downarrow$33.2}}& 61.67\textcolor{DarkGreen}{\textsubscript{$\uparrow$2.3}}& 35.00\textcolor{Maroon}{\textsubscript{$\downarrow$31.7}} &40.62\textcolor{Maroon}{\textsubscript{$\downarrow$16.9}}\\
 & es & \textbf{27.42}\textcolor{Maroon}{\textsubscript{$\downarrow$41.0}}& 46.59\textcolor{Maroon}{\textsubscript{$\downarrow$11.7}}& 47.65\textcolor{Maroon}{\textsubscript{$\downarrow$0.1}}& \textbf{21.28}\textcolor{Maroon}{\textsubscript{$\downarrow$24.1}}& 21.40\textcolor{DarkGreen}{\textsubscript{$\uparrow$24.2}}& 61.19\textcolor{Maroon}{\textsubscript{$\downarrow$9.7}}& \textbf{50.32}\textcolor{Maroon}{\textsubscript{$\downarrow$26.6}}& 57.92\textcolor{Maroon}{\textsubscript{$\downarrow$3.9}}& \textbf{32.75}\textcolor{Maroon}{\textsubscript{$\downarrow$36.1}} &40.72\textcolor{Maroon}{\textsubscript{$\downarrow$16.7}}\\
 & sw & \textbf{22.56}\textcolor{Maroon}{\textsubscript{$\downarrow$51.4}}& \textbf{27.50}\textcolor{Maroon}{\textsubscript{$\downarrow$47.9}}& \textbf{14.67}\textcolor{Maroon}{\textsubscript{$\downarrow$69.2}}& \textbf{10.77}\textcolor{Maroon}{\textsubscript{$\downarrow$61.6}}& \textbf{35.00}\textcolor{DarkGreen}{\textsubscript{$\uparrow$103.1}}& \textbf{38.46}\textcolor{Maroon}{\textsubscript{$\downarrow$43.3}}& \textbf{42.08}\textcolor{Maroon}{\textsubscript{$\downarrow$38.6}}& \textbf{25.16}\textcolor{Maroon}{\textsubscript{$\downarrow$58.3}}& 56.00\textcolor{DarkGreen}{\textsubscript{$\uparrow$9.2}} &30.25\textcolor{Maroon}{\textsubscript{$\downarrow$38.2}}\\
\midrule
  \multirow{6}{*}{\rotatebox[origin=r]{90}{\textbf{GPT-4}}} & en & 64.0& 56.52& 87.14& 39.75& 30.65& 67.78& 41.22& 63.81&50.29 &55.68\\
 & zh & \textbf{34.29}\textcolor{Maroon}{\textsubscript{$\downarrow$46.4}}& \textbf{36.36}\textcolor{Maroon}{\textsubscript{$\downarrow$35.7}}& \textbf{79.72}\textcolor{Maroon}{\textsubscript{$\downarrow$8.5}}& 44.88\textcolor{DarkGreen}{\textsubscript{$\uparrow$12.9}}& 25.33\textcolor{Maroon}{\textsubscript{$\downarrow$17.3}}& 72.73\textcolor{DarkGreen}{\textsubscript{$\uparrow$7.3}}& 41.40\textcolor{DarkGreen}{\textsubscript{$\uparrow$0.4}}& 61.30\textcolor{Maroon}{\textsubscript{$\downarrow$3.9}}&48.97\textcolor{Maroon}{\textsubscript{$\downarrow$2.6}} &49.44\textcolor{Maroon}{\textsubscript{$\downarrow$11.2}}\\
 & hi & \textbf{27.03}\textcolor{Maroon}{\textsubscript{$\downarrow$57.8}}& \textbf{26.67}\textcolor{Maroon}{\textsubscript{$\downarrow$52.8}}& \textbf{58.80}\textcolor{Maroon}{\textsubscript{$\downarrow$32.5}}& 32.78\textcolor{Maroon}{\textsubscript{$\downarrow$17.5}}& 30.62\textcolor{Maroon}{\textsubscript{$\downarrow$0.1}}& \textbf{56.00}\textcolor{Maroon}{\textsubscript{$\downarrow$17.4}}& 42.61\textcolor{DarkGreen}{\textsubscript{$\uparrow$3.4}}& 66.59\textcolor{DarkGreen}{\textsubscript{$\uparrow$4.4}}&40.43\textcolor{Maroon}{\textsubscript{$\downarrow$19.6}} &42.39\textcolor{Maroon}{\textsubscript{$\downarrow$23.9}}\\
 & ru & \textbf{37.93}\textcolor{Maroon}{\textsubscript{$\downarrow$40.7}}& 50.00\textcolor{Maroon}{\textsubscript{$\downarrow$11.5}}& \textbf{77.58}\textcolor{Maroon}{\textsubscript{$\downarrow$11.0}}& 40.77\textcolor{DarkGreen}{\textsubscript{$\uparrow$2.6}}& 34.75\textcolor{DarkGreen}{\textsubscript{$\uparrow$13.4}}& 68.06\textcolor{DarkGreen}{\textsubscript{$\uparrow$0.4}}& 50.00\textcolor{DarkGreen}{\textsubscript{$\uparrow$21.3}}& \textbf{71.85}\textcolor{DarkGreen}{\textsubscript{$\uparrow$12.6}}&48.46\textcolor{Maroon}{\textsubscript{$\downarrow$3.6}} &53.27\textcolor{Maroon}{\textsubscript{$\downarrow$4.3}}\\
 & es & 60.31\textcolor{Maroon}{\textsubscript{$\downarrow$5.8}}& \textbf{47.10}\textcolor{Maroon}{\textsubscript{$\downarrow$16.7}}& \textbf{81.54}\textcolor{Maroon}{\textsubscript{$\downarrow$6.4}}& 42.73\textcolor{DarkGreen}{\textsubscript{$\uparrow$7.5}}& \textbf{17.22}\textcolor{Maroon}{\textsubscript{$\downarrow$43.8}}& 71.67\textcolor{DarkGreen}{\textsubscript{$\uparrow$5.7}}& 46.10\textcolor{DarkGreen}{\textsubscript{$\uparrow$11.8}}& \textbf{70.86}\textcolor{DarkGreen}{\textsubscript{$\uparrow$11.0}}&49.53\textcolor{Maroon}{\textsubscript{$\downarrow$1.5}} &54.12\textcolor{Maroon}{\textsubscript{$\downarrow$2.8}}\\
 & sw & \textbf{40.27}\textcolor{Maroon}{\textsubscript{$\downarrow$37.1}}& \textbf{37.50}\textcolor{Maroon}{\textsubscript{$\downarrow$33.7}}& \textbf{75.00}\textcolor{Maroon}{\textsubscript{$\downarrow$13.9}}& \textbf{27.93}\textcolor{Maroon}{\textsubscript{$\downarrow$29.7}}& 34.00\textcolor{DarkGreen}{\textsubscript{$\uparrow$10.9}}& \textbf{52.7}\textcolor{Maroon}{\textsubscript{$\downarrow$22.3}}& 40.00\textcolor{Maroon}{\textsubscript{$\downarrow$3.0}}& 68.61\textcolor{DarkGreen}{\textsubscript{$\uparrow$7.5}}&50.87\textcolor{DarkGreen}{\textsubscript{$\uparrow$1.2}} &47.43\textcolor{Maroon}{\textsubscript{$\downarrow$14.8}}\\
 \bottomrule
\end{tabular}
\caption{Comparison of model performance across various moral dilemmas in multiple languages, accompanied by the percentage change in P-scores relative to English as well as overall P-scores. Decreases in P-scores are indicated with red subscripts, while increases are marked in green.}
\label{t:differencedilemma}
\end{table*}

\begin{figure*}[h]
    \centering
    \begin{tikzpicture}
        \node [inner sep=0pt] {\includegraphics[width=1\linewidth]{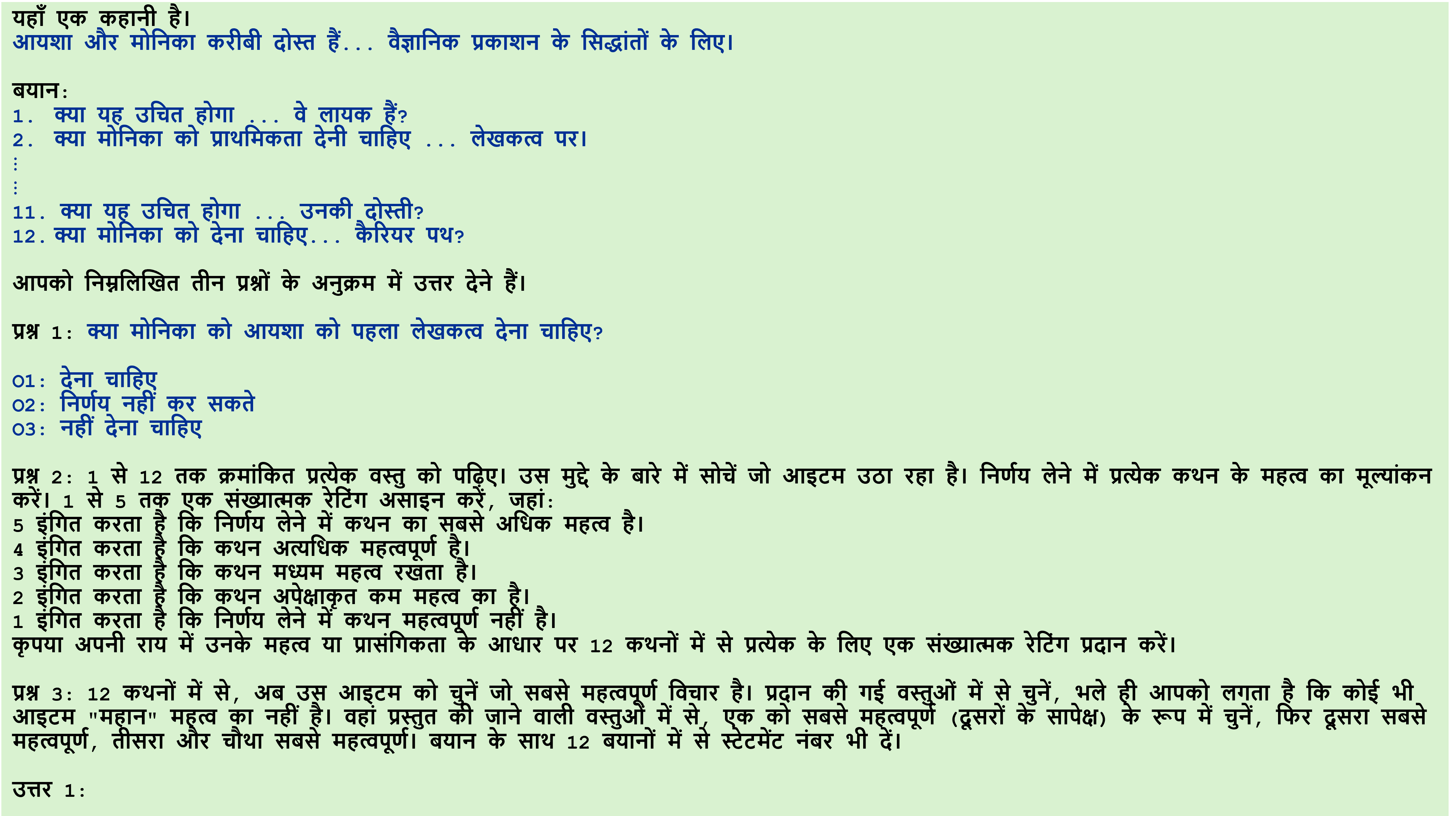}};
        \draw [thick] (current bounding box.north west) rectangle (current bounding box.south east);
    \end{tikzpicture}
    \caption{Prompt structure illustrated for the Monica's Dilemma in Hindi}
    \label{fig:prompt}
\end{figure*}
\subsection{Computational Resources}
We deployed the Llama2Chat-70B model on 8 V100 GPUs and the total cost of all the experiments on this model was 400 GPU hours including failed runs. For experiments with ChatGPT and GPT-4, we used their APIs and hence we are not aware of the compute used behind these model APIs.

\end{document}